\documentclass[pdflatex,sn-apa,iicol]{sn-jnl}

\usepackage{graphicx}%
\usepackage{multirow}%
\usepackage{amsmath,amssymb,amsfonts}%
\usepackage{amsthm}%
\usepackage{mathrsfs}%
\usepackage[title]{appendix}%
\usepackage{xcolor}%
\usepackage{textcomp}%
\usepackage{manyfoot}%
\usepackage{booktabs}%
\usepackage{algorithm}%
\usepackage{algorithmicx}%
\usepackage{algpseudocode}%
\usepackage{listings}%

\usepackage{booktabs}
\usepackage{tabularx}
\usepackage{array}
\usepackage{makecell}

\usepackage{amssymb}
\usepackage{amsmath}
\usepackage{booktabs}
\usepackage{pifont}
\usepackage{lineno,hyperref}

\theoremstyle{thmstyleone}%
\theoremstyle{thmstyletwo}%

\theoremstyle{thmstylethree}%

\begin{document}

\title[Article Title]{EHGCN: Hierarchical Euclidean-Hyperbolic Fusion via Motion-Aware GCN for Hybrid Event Stream Perception}









         

\author[1]{Haosheng Chen}\email{chenhs@cqupt.edu.cn}
\author[1]{Lian Luo}\email{s240231064@stu.cqupt.edu.cn}
\author[1]{Mengjingcheng Mo}\email{mo1031@live.com}
\author[1]{Zhanjie Wu}\email{s230201119@stu.cqupt.edu.cn}
\author[1]{Ji Gan}\email{ganji@cqupt.edu.cn}
\author[1]{\quad Jiaxu Leng}\email{lengjx@cqupt.edu.cn}
\author*[1]{Xinbo Gao}\email{gaoxb@cqupt.edu.cn}

\affil[1]{Chongqing Key Laboratory of Image Cognition, Chongqing University of Posts and Telecommunications, Chongqing 400065, China}








\abstract{Event cameras, characterized by microsecond temporal resolution and very High Dynamic Range (HDR), emit high-speed event streams for perception tasks. In recent advancements, Graph Neural Networks (GNNs)-based methods show great potential in event perception. However, they typically rely on  straightforward pairwise node connectivity in Euclidean space where they struggle to capture long-range dependencies and faithfully characterize the inherent hierarchical structures of event streams. To this end, we propose EHGCN, a dual-space event perception approach that, to the best of our knowledge, is the first to jointly model event streams in Euclidean and hyperbolic spaces. By introducing hyperbolic geometry into event stream perception, EHGCN enables to naturally capture the anisotropic and hierarchical structures of non-uniform, motion-driven event streams. Specifically, we first introduce a distribution-aware event sifting method based on multi-scale voxel grids and Gaussian distribution modeling, retaining discriminative events while attenuating chaotic noise. Then, we present a Markov Random Field (MRF)-optimized motion-aware hyperedge generation scheme, which minimizes a motion consistency energy function to explicitly capture consistent global motion patterns within short time intervals, thereby eliminating cross-target spurious associations and providing critically topological priors while capturing long-range dependencies among events. Finally, we propose a Euclidean-hyperbolic GCN to fuse the retinal events densely aggregated and hierarchically modeled in local Euclidean and global hyperbolic spaces, respectively, to achieve a hybrid event perception. Extensive experimental results on event perception tasks, such as object detection and recognition, show the effectiveness of our approach. Our code will be released for public use at https://github.com/ev-lluo/EHGCN.}




\maketitle

\section{Introduction}
\begin{figure*}[t]
  \centering
   \includegraphics[width=1\linewidth]{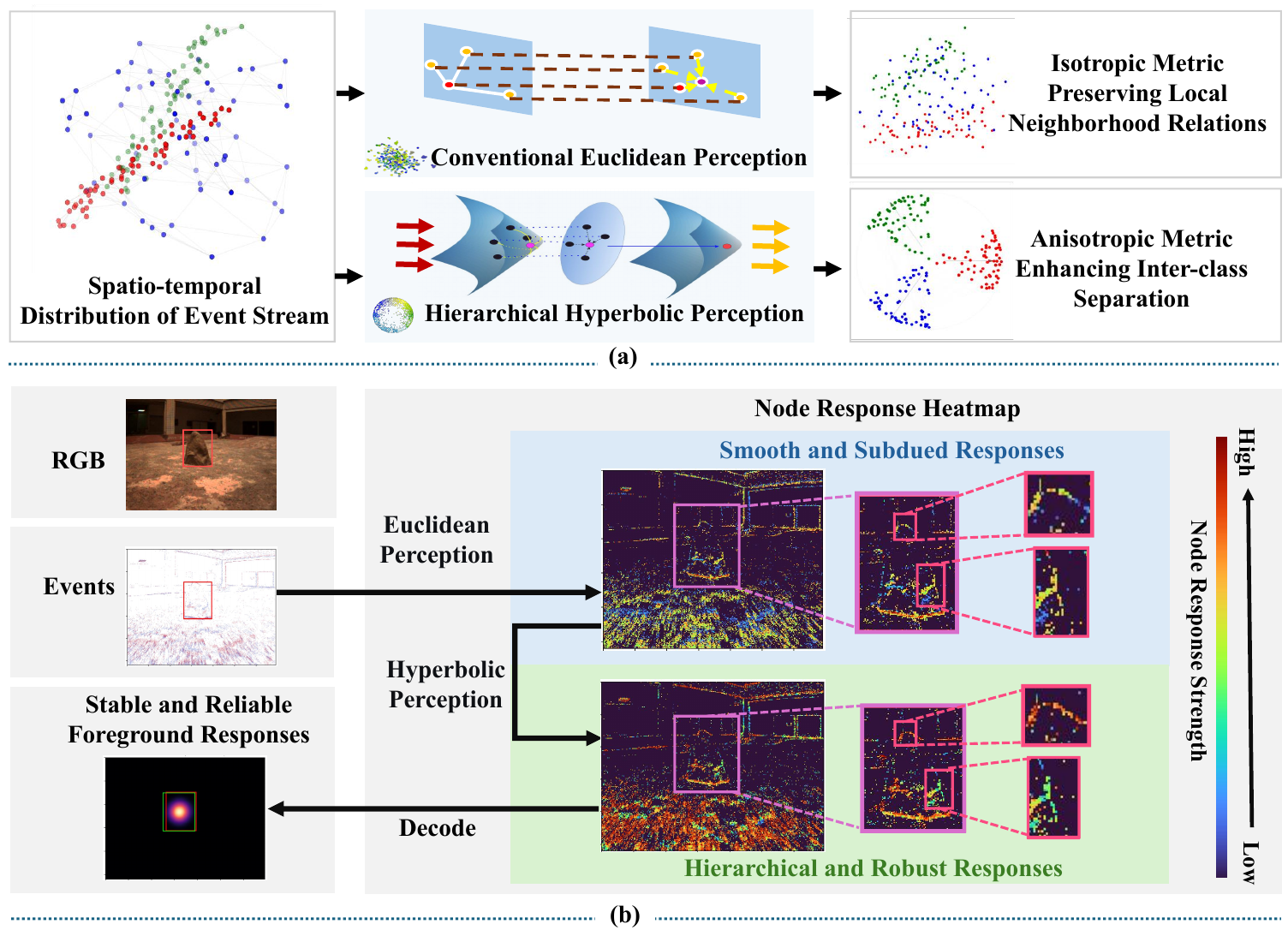}
   \caption{Euclidean and hyperbolic spaces offer complementary representational strengths. (a) shows that event streams exhibit global spatio-temporal anisotropy and hierarchical organization while preserving strong local spatio-temporal continuity. Conventional Euclidean geometry enables to exploit an isotropic metric to preserve local neighborhood relations, while hyperbolic geometry can leverage an anisotropic metric to enhance inter-class separation. (b) further illustrates the node response differences. Euclidean perception produces relatively smooth and subdued responses, while hyperbolic perception keeps the main response regions spatially stable and reorganizes response intensities into clearer hierarchical patterns. This enables the model to better perceive foreground objects, understand complex background structures, and capture robust features such as corners, thereby contributing to more stable and reliable detection responses.}
   \label{image:8}
\end{figure*}

Event cameras have gained significant attention in computer vision due to their ability to asynchronously record brightness changes at each pixel. This mechanism emulates the functional characteristics of retinal ganglion cells, enabling microsecond-level responsiveness and low data redundancy \citep{ijcv4_deraining}. Thus, it demonstrates remarkable advantages for tasks in event-based perception such as object detection \citep{re_detecctionCNN}, recognition \citep{2026IJCV_human_action_recognition} and tracking \citep{2026IJCV_tracking}. Nevertheless, the inherent sparse and asynchronous characteristics of event streams pose significant challenges on perception tasks.

For perception, traditional frame-based methods convert event streams into dense tensors \citep{re_timesurfaces} using fixed temporal windows or adaptive spatio-temporal aggregation, misaligning with the inherent sparsity and asynchronicity of event data, causing information loss and computational redundancy. Thus, direct processing of event streams is paramount. While Spiking Neural Networks (SNNs) are naturally compatible with event data \citep{introduction4_recognition}, their hardware parallelism demands and suboptimal learning algorithms hinder practical deployment. Geometric graph representations of Graph Neural Networks (GNNs) address these issues by modeling events as graph nodes and constructing edges based on spatio-temporal proximity, forming structured spatio-temporal event graphs. This paradigm preserves sparsity while enabling topological analysis, demonstrating competitive performance across tasks \citep{EAGR_recog}. However, since a moving object exhibits unique motion pattern over a very short time interval, it generates numerous events that share the same motion state. Traditional spatio-temporal graphs treat these events as graph nodes connected by pairwise edges. While this paradigm can model local interactions, it struggles to directly capture the shared motion state among global events, resulting in redundant representations and structural limitations, which limits their ability to preserve event stream continuity and capture long-range spatio-temporal dependencies.

Furthermore, current GNN-based methods perform metric learning and feature aggregation within Euclidean space. While Euclidean space excels at capturing local correlations, its isotropic metric nature makes it difficult to effectively characterize the intrinsic hierarchical structure induced by motion differences in event streams. Objects with different motion patterns form tight yet distinct spatio-temporal clusters of events over short intervals, whereas low-speed backgrounds or noise correspond to sparse and scattered event distributions. This spatio-temporal anisotropy results in a naturally hierarchical structure in the event data. Due to the limited volume growth in Euclidean space, it is prone to significant representation distortion when handling such complex hierarchical structures. In contrast, hyperbolic space \citep{poincare}, with negative curvature, can embed this \textit{time-dominated and directionally prominent} structure with lower distortion through non-uniform scaling and exponential volume growth \citep{2026hyperbolic_1}, offering a natural advantage in hierarchical modeling and long-range dependency capture. Moreover, the strong representation capacity of hyperbolic geometry allows hierarchical structures to be encoded in a more compact embedding space, which can reduce the dependence on high-dimensional Euclidean embeddings, excessive channel expansion, and redundant deep transformations. Therefore, hyperbolic modeling provides a parameter-efficient way to learn expressive event representations, enabling the model to capture complex spatio-temporal hierarchies with fewer learnable parameters while preserving discriminative structural information.

Euclidean and hyperbolic spaces offer complementary representational strengths. As shown in Fig. \ref{image:8}(a), although event streams globally exhibit spatio-temporal anisotropy and hierarchical organization due to their non-uniform and motion-driven nature, they still maintain strong local spatio-temporal continuity within neighborhoods. Conventional Euclidean perception can employ the isotropic metric to preserve local neighborhood relations for effective local structural modeling, whereas hierarchical hyperbolic perception can leverage the anisotropic metric to enhance inter-class separation and better represent the hierarchical organization of event streams. As shown in Fig. \ref{image:8}(b), after Euclidean GCN processing, the node responses are mainly distributed within low and medium intensity ranges, exhibiting an overall smooth response pattern. Such distribution can preserve the major spatial structures, but their ability to hierarchically distinguish foreground objects, complex backgrounds, and local structural cues remains limited. After further Hyperbolic GCN processing, the principal response regions remain spatially stable, while the response intensities are reorganized into clearer low, medium, and high levels. This hierarchical response pattern enables the model to simultaneously learn and perceive foreground objects and complex background structures, and further understand more robust and trustworthy scene features, such as the corner features shown in the figure, thereby helping to produce stable and reliable detection responses. Therefore, Euclidean GCN helps preserve local intra-class geometric structures, while Hyperbolic GCN further enhances global hierarchical relationships and inter-class separability.

Given the above insights, we formulate a Euclidean-Hyperbolic hybrid perception model for event stream. Considering the computational burden brought by the large amount of event data and the noise interference caused by environmental and sensor factors, we propose a Distribution-Aware Event Sifting (DAES) method that constructs local Gaussian models within multi-scale voxel grids, which can preserve discriminative events and attenuate chaotic noise. Then, based on the preliminary estimation of event velocity, we introduce a Markov Random Field (MRF) \citep{IF003} to further refine the estimation. By leveraging the inherent spatio-temporal continuity and motion coherence in event data, hyperedges are generated to capture group-wise interactions beyond pairwise connections. To leverage the complementary advantages of both Euclidean and hyperbolic spaces, we present a hierarchical dual-space fusion where Euclidean GCN models local event interactions to preserve intra-class geometric structures, while hyperbolic GCN captures global hierarchical dependencies to enhance inter-class separability.

To sufficiently evaluate the effectiveness and robustness of our model in complex event-based perception scenarios, we expect the model to perform perception tasks across diverse environments. However, existing event-based datasets are mainly concentrated on traffic scenarios \citep{gen1,ncars} or general indoor \citep{dvs128} and outdoor environments \citep{ijcv2025_urban_dataset}. Therefore, we construct EVMars-Detection for planetary exploration scenarios, focusing on rock localization and fine-grained category discrimination by Mars rovers. The dataset covers multiple representative rock categories, different rover motion speeds, and complex acquisition conditions such as dust disturbance, while also incorporating practical challenges including occlusion, cluttered backgrounds, and HDR illumination. EVMars-Detection provides a more targeted and challenging benchmark for evaluating the stable perception capability of our models in complex extraterrestrial environments.

In summary, our contributions can be summarized as follows:

\begin{itemize}
    
    \item \textbf{Hyperbolic Event Embedding:} The first to introduce retinal events into the hyperbolic space. It captures hierarchy of the multi-scale motion states and represents the global motion consistency across spaces and times.
    \item \textbf{Distribution-Aware Event Sifting:} A distribution-aware method that constructs local Gaussian models within multi-scale voxel grids and adaptively suppresses unreliable events according to distribution consistency, preserving discriminative event structures while attenuating chaotic noise.
    \item \textbf{Motion-Aware Hypergraph Modeling:} A MRF-optimized motion-aware hypergraph generation scheme is proposed to generate hyperedges with spatio-temporal continuity. It optimizes motion estimation by minimizing an energy function, explicitly capturing consistent motion patterns within short time intervals.
    \item \textbf{Dual-Space Hybrid Fusion: }A dual-space hybrid learning paradigm, in which the Euclidean space with an isotropic metric preserves local intra-class geometric structures via neighborhood interactions, while the hyperbolic space leverages its negatively curved geometry to accommodate anisotropic hierarchies and model global relationships, thereby effectively improving inter-class separability.
    \item \textbf{EVMars-Detection Dataset:} A multi-modal dataset for event-based Mars-like terrain detection, featuring diverse motion patterns and severe challenges such as dust disturbances, occlusions, cluttered backgrounds, and HDR illumination. It provides a tough benchmark for evaluating event-based perception under complex extraterrestrial conditions.
\end{itemize}

\section{Related Works}
\subsection{Event Data Representation}

 Event streams are inherently sparse and asynchronous, while exhibiting strong spatio-temporal correlations, posing substantial challenges for effective representation. Accordingly, current event representation methods can be broadly grouped into two paradigms, each associated with a corresponding processing architecture, as described below:

\textbf{Frame-based representation methods.} These research predominantly focuses on accumulating asynchronous event streams into regular, image-like frames (e.g., event histogram \citep{mamba}, time surfaces \citep{ijcv3_Time-Surface}, or voxel grids \citep{re_voxelgrids}) through fixed temporal windows or adaptive spatio-temporal aggregation \citep{2026adaptive_Spatial-Temporal_Window_3}. This discretization operation converts the event streams into dense tensors, making them compatible with conventional Convolutional Neural Networks (CNNs) architectures for object detection \citep{re_detecctionCNN} or classification \citep{re_classification_CNN}. With the development of computer vision, due to their larger receptive fields, Transformer-based \citep{re_transformer2} methods have significantly shown superior performance over CNN-based methods. More recently, inspired by state-space model (SSM), Mamba \citep{mamba} proposes a time-varying state-space model equipped with a selection mechanism. It successfully models long sequences effectively with complexity scaling linearly with sequence length, achieving a better balance between efficiency and performance, and having emerged as a promising alternative to Transformers \citep{re_mamba2}. However, these methods neglect the inherent sparsity and asynchronicity of the event stream due to their frame-based representation scheme, leading to information loss and computational redundancy.

\textbf{Point-based representation methods.} Characterized by their ability to efficiently process the raw event streams in a sparse and asynchronous manner. Dominant architectures include SNN \citep{re_SNN_recognition} and GNN \citep{NVS-S,RGCNNS_recog}. SNN, as neuromorphic models, align with event cameras through event-driven processing and temporal precision. However, their strict hardware parallelism and suboptimal learning algorithms hinder deployment. Compared with SNN, graph-based geometric learning provides a compelling approach for event data representation. These representations construct spatio-temporal event graphs by treating events as nodes connected through spatio-temporal proximity, modeling relationships via asynchronous mechanisms. While GNNs have made progress in tasks like object detection \citep{aegnn}, recognition \citep{NVS-S} and tracking \citep{re_GNN_tracking}, traditional graph networks still face critical limitations, since the current methods mainly focus on pairwise relationships, failing to capture global dependencies and complex and hierarchical dynamic patterns. 

Hypergraphs, unlike traditional graphs, can use hyperedges to connect multiple nodes \citep{ijcv1_Hyper,2025hypergraph_1}, capturing complex high-order interactions \citep{2026softhgnn}. Hypergraphs have achieved promising performance in various computer vision tasks, including pose estimation \citep{re_hyper_poseestimation1}, action recognition \citep{eventhypergraph}, and image classification and detection \citep{vihgnn}, demonstrating their effectiveness in capturing complex relationships and high-dimensional structures. These studies provide valuable insights for modeling high-order event relationships beyond local pairwise interactions in event streams.

\subsection{Hyperbolic Geometric Learning}
Hyperbolic geometry with constant negative curvature excels at modeling hierarchical and high‑dimensional nonlinear relationships in visual hierarchies \citep{ijcv2_Hyperbolic,2026balanced}, enabling low‑distortion embedding of exponentially growing patterns into compact spatial domains and outperforming Euclidean space in preserving hierarchical geometric properties \citep{poincare}. This geometric superiority has driven hyperbolic graph neural network (HGNN) to demonstrate exceptional performance in recommender systems \citep{re_bolic_recommendersystems3} and knowledge graphs \citep{re_bolic_knowledgegraphs3}. And their applications are expanding into computer vision core tasks like few-shot object recognition \citep{re_bolic_recognition} and object detection \citep{re_bolic_detection}. Notably, \citet{introductionHGCN_re_bolic} first merge hyperbolic manifolds with GCNs in HGCN, effectively overcoming Euclidean space limits in hierarchical graph processing.

\begin{figure*}[t]
  \centering
   \includegraphics[width=1\linewidth]{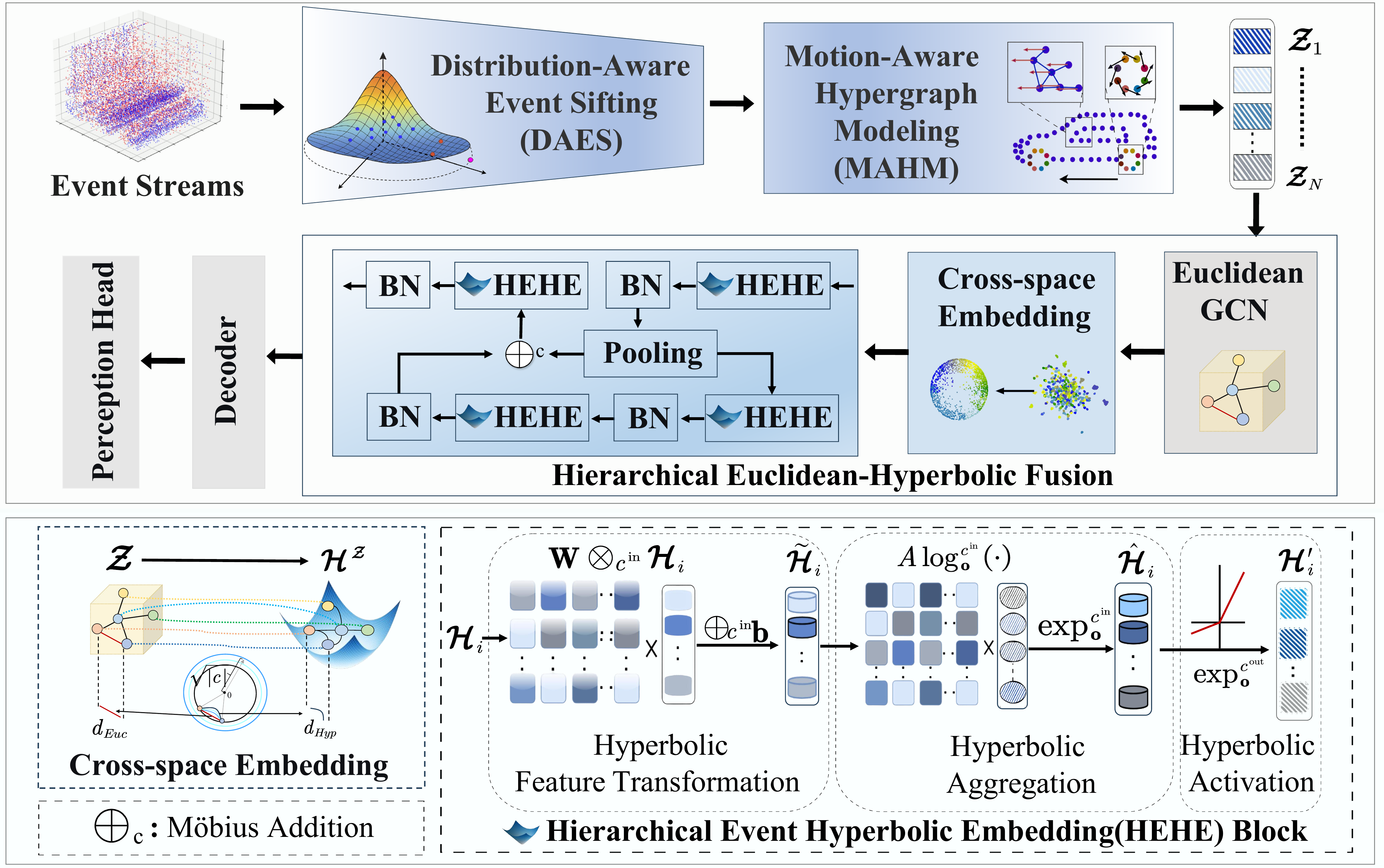}
   \caption{Framework of the proposed EHGCN approach. Our distribution-aware event sifting method models local event distributions within multi-scale voxel grids by performing Gaussian modeling, preserving discriminative events while attenuating chaotic noise. Then EHGCN generates motion-aware hypergraphs through MRF-optimized estimation, capturing long-range spatio-temporal dependencies and motion coherence in event streams. After that, EHGCN fuses the information densely aggregated and hierarchically modeled in local Euclidean and global hyperbolic spaces, respectively.}
   \label{image:1}
\end{figure*}

In addition, our previous studies have demonstrated the strong capability of hyperbolic space for complex visual relation modeling. For weakly supervised video violence detection, we introduced hyperbolic representations to improve the discrimination of visually similar but semantically different violent and non-violent events \citep{wuzhanjie1}. In vision-language guided violence detection, hyperbolic space was further utilized to construct hierarchical semantic constraints for recognizing confusing samples in complex scenarios \citep{wuzhanjie2}. For video person re-identification, we proposed dual-space representation learning to jointly capture local appearance features and hierarchical identity relationships, improving cross-camera and cross-time matching robustness \citep{kuangChangjiang}. Moreover, hyperbolic space was incorporated into normalizing flow modeling for unsupervised video anomaly detection, enabling better characterization of hierarchical differences among similar skeleton actions \citep{zhangyumeng}. These studies demonstrate that hyperbolic space effectively complements Euclidean representations in modeling hierarchical relationships. However, existing hyperbolic vision methods mainly focus on conventional data, while their potential for event-based perception remains largely unexplored.

Drawing inspiration from the aforementioned insights, we propose EHGCN, taking the advantages of hypergraphs and hyperbolic geometry. To the best of our knowledge, this is the first introduction of hyperbolic space for perceiving event-based data. This approach leverages hypergraphs to capture complex high-order interactions and exploits hyperbolic geometry to effectively model hierarchical and nonlinear relationships in event streams, addressing the critical limitations of conventional Euclidean methods.

\section{Preliminaries}

\textbf{Hyperbolic Geometry.} Hyperbolic geometry, with constant negative curvature, enables exponential expansion of hierarchical structures, providing unique advantages for modeling multi-scale motion patterns in event streams. We adopt the Poincaré ball model \citep{introductionHGCN_re_bolic} to define an n-dimensional hyperbolic space $\small\mathrm{D}_{c}^{n}$, where the negative curvature magnitude $\small\ -c$ is controlled by a curvature $\small\ c>0$. The Poincaré ball maps hyperbolic space to a unit ball in Euclidean space: $\small\mathrm{D}_{c}^{n}=\left\{ \mathbf{x}\in \mathrm{R}^n\left| \,\,\left\| \mathbf{x} \right\| ^2<\frac{1}{c} \right. \right\} $, with the Riemannian metric defined as: $\small\ g^0=\lambda _{c}^{2}g^E$, where $\small\ g^E=\mathbf{I}_n$ is the Euclidean metric tensor and $\small\lambda _c=\frac{2}{1-c\left\| \mathbf{x} \right\| ^2}$ is the conformal factor. 

\textbf{Tangent Space.} Unlike Euclidean space, hyperbolic space is not a vector space. Therefore, linear operations such as Euclidean distance computation or vector addition cannot be directly performed in hyperbolic space and need to be conducted with the aid of tangent spaces. For a point $\small\mathbf{x}\in \mathrm{D}_{c}^{n}$, the tangent space $\small\mathbf{T}_{\mathbf{x}}\mathrm{D}_{c}^{n}$ describes the set of tangent vectors through it. A tangent vector $\small\mathbf{u}\in \mathbf{T}_{\mathbf{x}}\mathrm{D}_{c}^{n}$ must satisfy the orthogonality condition: $\small\left< \mathbf{u},\mathbf{x} \right> _{\mathrm{D}_{c}^{n}}=0$, indicating that $\mathbf{u}$  is orthogonal to the radial direction from the origin to $\small\mathbf{x}$.

\textbf{Hyperbolic Distance and Inner Product.} For two points $\small\mathbf{x},\mathbf{y}\in \mathrm{D}_{c}^{n}$ the hyperbolic distance and inner product are defined as: 
\begin{equation}
  \small\mathbf{d}_{\mathrm{D}_{c}^{n}}(\mathbf{x},\mathbf{y})=\frac{1}{\sqrt{c}}\mathrm{arcosh}\mathrm{(}1+\frac{2c\left\| \mathbf{x}-\mathbf{y} \right\| ^2}{\left( 1-c\left\| \mathbf{x} \right\| ^2 \right) \left( 1-c\left\| \mathbf{y} \right\| ^2 \right)}),
\end{equation}
\begin{equation}
  \small\left< \mathbf{x},\mathbf{y} \right> _{\mathrm{D}_{c}^{n}}=-\mathbf{x}_0\mathbf{y}_0+c\sum_{i=1}^n{\mathbf{x}_i\mathbf{y}_i}.
\end{equation}

The hyperbolic distance explicitly models nonlinear relationships in negatively curved spaces, where distances contract with increasing curvature. The inner product serves to measure the similarity of the features between the nodes.

\begin{figure*}[t]
  \centering
   \includegraphics[width=1\linewidth]{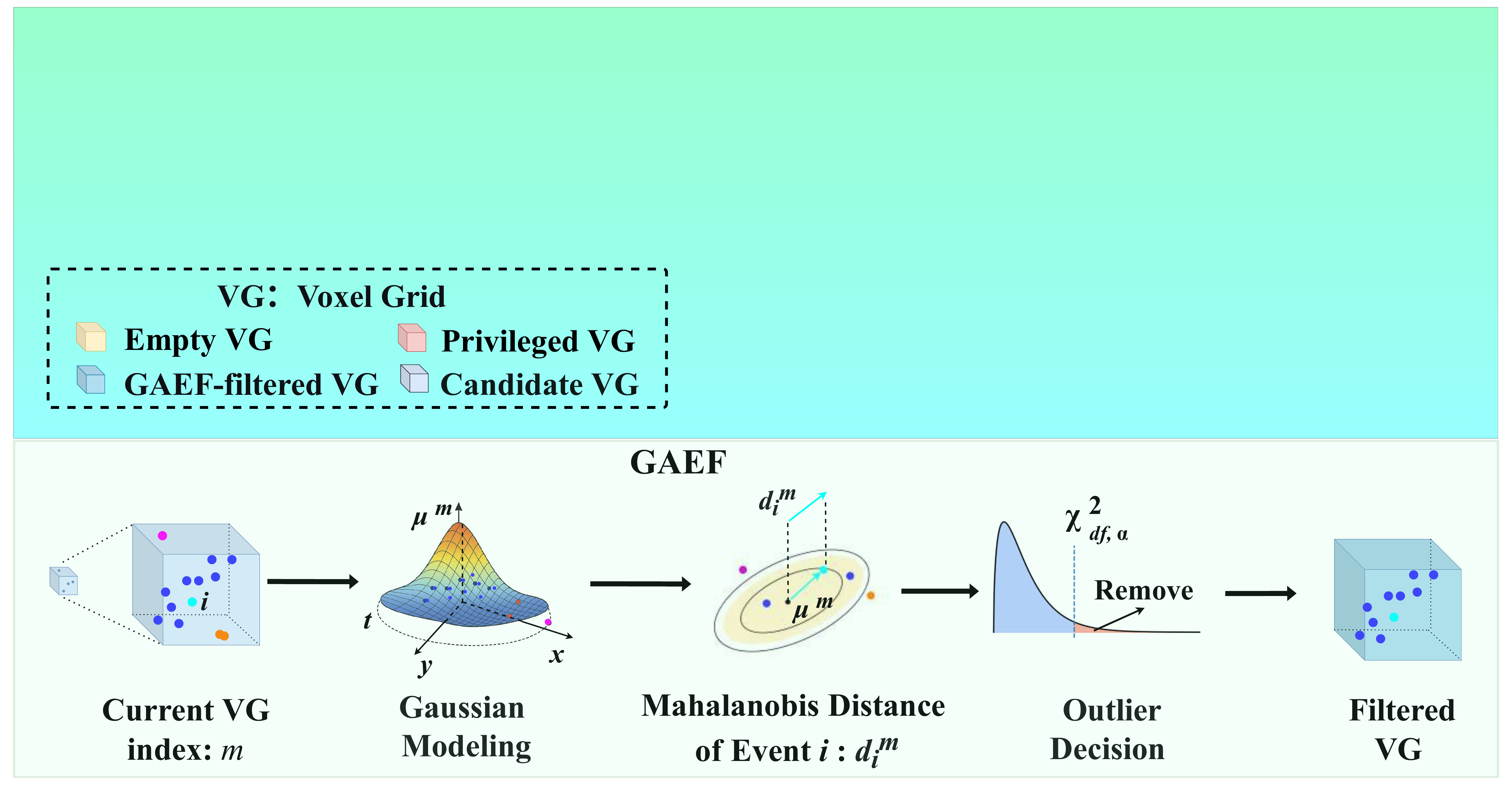}
  
   \caption{The proposed DAES. We performs Gaussian modeling of events within multi-scale 3D voxel grids and adaptively removes outliers based on the Mahalanobis distance, preserving discriminative events while attenuating chaotic noise.
   }
   \label{image:2}
\end{figure*}

\textbf{Möbius Addition.} The Möbius addition in hyperbolic space is defined as:
\begin{equation}
    \small\mathbf{x}\oplus _c\mathbf{y}=\frac{\left(1+2c\left< \mathbf{x},\mathbf{y}\right> _{\mathrm{D}_{c}^{n}}+c\left\| \mathbf{y} \right\| ^2 \right) \mathbf{x}+\left( 1-c\left\| \mathbf{x} \right\| ^2 \right) \mathbf{y}}{1+2c\left< \mathbf{x},\mathbf{y} \right> _{\mathrm{D}_{c}^{n}}+c^2\left\| \mathbf{x} \right\| ^2\left\| \mathbf{y} \right\| ^2},
\end{equation}
which preserves the structural invariance of hyperbolic space and is suitable for cross-hierarchical feature aggregation.

\textbf{Möbius Matrix-Vector Multiplication.} The extension of linear transformations to hyperbolic space is formulated as:
\begin{equation}
   \small W\otimes _c\mathbf{x}=\exp _{\mathbf{o}}^{\mathrm{c}}\left( W\log _{\mathbf{o}}^{\mathrm{c}}\left( \mathbf{x} \right) \right), 
\end{equation}
where $\small\ W$ is a learnable weight matrix. $\small\exp _{\mathbf{o}}^{\mathrm{c}}$ and $\small\log _{\mathbf{o}}^{\mathrm{c}}$ denote the exponential and logarithmic maps anchored at the origin respectively.

\section{Methodology}   
\label{Sec:4}

\textbf{Overview.} EHGCN proposes Distribution-Aware Event Sifting (DAES), Motion‑Aware Hypergraph Modeling (MAHM) and dual‑space fusion architecture for robust, hierarchy‑aware event perception. As shown in Fig. \ref{image:1}, EHGCN first constructs Gaussian distribution models to characterize local event distributions within multi-scale voxel grids and adaptively suppresses unreliable noisy events according to their consistency with the corresponding local distributions. It then constructs motion‑consistent hyperedges through MRF-optimized motion estimation. After that, a Euclidean GCN is applied to model local interactions and preserve the clarity of intra-class structures, followed by a hyperbolic GCN to enhance inter-class separation and better capture global spatio-temporal dependencies. This dual-space fusion architecture effectively balances local geometric properties and global hierarchical topological properties.

\subsection{Distribution-Aware Event Sifting}
Traditional cameras capture frames at a fixed frame rate, synchronously accumulating photon counts across all pixels during a predefined shutter interval. In contrast, event cameras asynchronously record logarithmic brightness changes \citep{ijcv5_RGB-D} at individual pixels, generating event streams containing pixel coordinates, timestamp, and polarity. An event ${\small\ e_k=\left( x_k,y_k,t_k,p_k \right)}$, where $\small\ p_k\in \left( -1,+1 \right)$ is the polarity indicating an increase or decrease in brightness, is triggered at pixel $\small\ X_k=(x_k,y_k)$ when the logarithmic brightness change exceeds the event threshold. A set of event streams within a time window $\Delta T$ is represented as an ordered tuple list $\small\varepsilon =\left\{ e_i \right\} _{i=1}^{N}$, where $ N $ is the number of events output by the event camera within $\Delta T$. In non-uniform event streams, object boundaries and high-speed motion regions typically exhibit dense event clusters with coherent spatio-temporal distribution patterns, while background regions mainly contain sparse and unreliable responses often caused by chaotic noise. Conventional global uniform sampling strategies ignore the intrinsic distribution characteristics of event streams. They may discard informative events in high-density regions while preserving unreliable responses in sparse regions, resulting in degraded event representation quality. 

Therefore, as shown in Fig. \ref{image:2}, we propose a Distribution-Aware Event Sifting (DAES) method to progressively select reliable events by modeling local event distribution characteristics. Specifically, the proposed method constructs Gaussian distribution models to characterize local events within multi-scale 3D voxel grids. The mean vector is used to describe the dominant spatio-temporal motion patterns of local events, while the covariance matrix characterizes the variation range of event distributions and spatial-temporal correlations. Based on this formulation, the Mahalanobis distance is adopted to measure the consistency between each event and its corresponding local distribution. Events with large deviations from local distributions are considered to have lower reliability and are adaptively removed, while events consistent with their local distributions are retained as reliable event representations.

\begin{figure*}[t]
  \centering
   \includegraphics[width=1\linewidth]{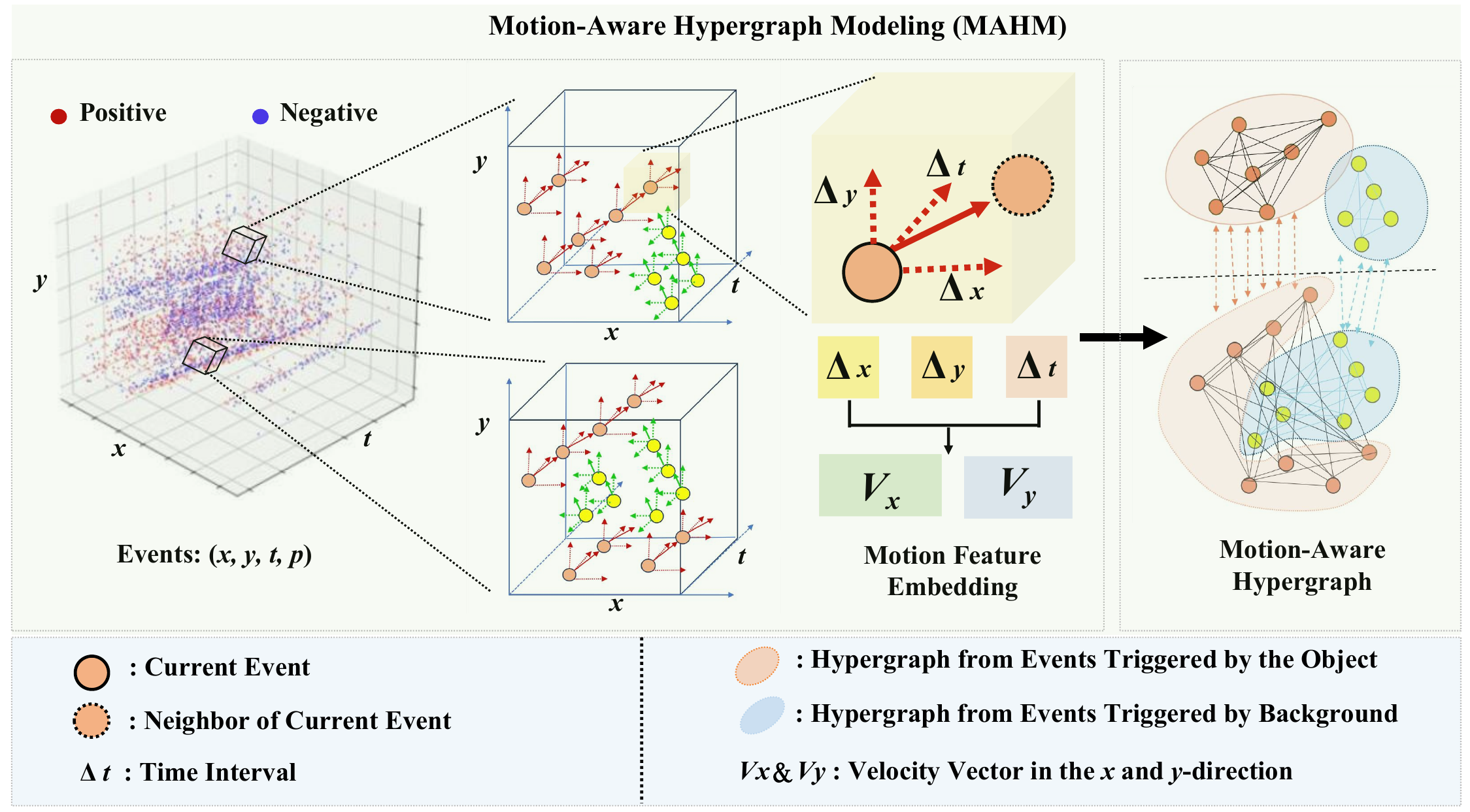}

   \caption{Motion-aware hypergraph. Events with similar motion patterns, corresponding to points of the same color, such as yellow and orange. The motion-aware hypergraph groups events with similar velocity patterns into hyperedges, enabling the modeling of high-order spatio-temporal interactions among events.}
    
   \label{image:3}
\end{figure*}

Let $P_i = (x_i, y_i, t_i)^\top \in \mathbb{R}^3$ denote the spatio-temporal position of event $i$, for $i = 1,\dots,N$. Given the voxel grid size $\mathbf{VG} = (VG_x, VG_y, VG_t)^\top$, the voxel index for an event at position $P_i$ is derived as:
\begin{equation}
    \mathbf{Voxel}(i) = \Bigl(\Bigl\lfloor \tfrac{x_i}{VG_x} \Bigr\rfloor, 
                  \Bigl\lfloor \tfrac{y_i}{VG_y} \Bigr\rfloor, 
                  \Bigl\lfloor \tfrac{t_i}{VG_t} \Bigr\rfloor\Bigr)^\top.
\end{equation}
For a non-empty candidate voxel grid $m$ with a set of event indices $\mathcal{I}_m = \{\,i \mid \mathbf{Voxel}(i)=m\,\}$ and $c_m = |\mathcal{I}_m|$. Next, we split $P_i$ to obtain $P_{i}^{s} = (x_i, y_i)^\top \in \mathbb{R}^2$ and $P_{i}^{t} = t_i$, where $i \in  \mathcal{I}_m$. Then we can calculate the spatial mean $\mu _{_{P^s}}^{m}$, temporal mean $\mu _{_{P^t}}^{m}$, and temporal standard deviation $\sigma _{_{P^t}}^{m}$ separately. Additionally, the spatial covariance can be obtained by:

\begin{equation}
    \small \Sigma _{_{P^s}}^{m}
= \frac{1}{c_m - 1}
\sum_{i \in I_m}
(P^s_i - \mu _{_{P^s}}^{m})
(P^s_i - \mu _{_{P^s}}^{m})^\top.
\end{equation}
From this, we can get the joint mean vector $\mu ^{m}=\left( \mu _{_{P^s}}^{m}, \mu _{_{P^t}}^{m} \right)^\top \in \mathbb{R}^{3 \times 3}$ and the joint covariance matrix $\Sigma^{m}$.
\begin{equation}
    \Sigma^{m} =
\begin{pmatrix}
\Sigma _{_{P^s}}^{m} & \mathbf{0}_{2 \times 1} \\
\mathbf{0}_{1 \times 2} & \bigl(\sigma _{_{P^t}}^{m})^2
\end{pmatrix}
\in \mathbb{R}^{3 \times 3}.
\end{equation}
Subsequently, we can calculate the Mahalanobis distance $d_i^{m}$.
\begin{equation}
    d_i^{m} = 
\sqrt{ \bigl(P_i - \boldsymbol{\mu}^{m}\bigr)^\top
\bigl(\Sigma^{m} )^{-1}
\bigl(P_i - \boldsymbol{\mu}^{m}\bigr) }.
\end{equation}
When the Mahalanobis distance of an event exceeds the boundary defined by the chi-square distribution $\chi^2_{df=3}$, it is considered an outlier and removed. The magnitude of the Mahalanobis distance directly reflects the typicality or anomaly of a point relative to the overall data population. The Mahalanobis distance incorporates the overall shape of the data distribution through the inverse covariance matrix. When there are correlations between dimensions, it automatically performs a relative spatio-temporal rotation and scaling, making the distance measure better match the true distribution. For event streams, this property is particularly suitable, as it can effectively identify and remove sparse, isolated outlier events.

As shown in Fig. \ref{image:2}, we perform the aforementioned Gaussian distribution modeling across voxel grids at different scales. This multi-scale representation enables progressive event reliability assessment from coarse to fine by capturing global distribution characteristics and local spatio-temporal details. Specifically, smaller voxel scales provide finer resolution for identifying subtle and highly localized event patterns, whereas larger voxel scales characterize broader distribution trends and suppress sparse unreliable responses caused by incidental noise. Compared with single-scale scheme, the proposed multi-scale sifting method validates events from complementary spatial-temporal perspectives and progressively filters unreliable responses while preserving informative and discriminative event structures in complex scenes.

\subsection{Motion-Aware Hypergraph Modeling}

Due to the temporal continuity of motion, an object's trajectory can be locally approximated by its first-order temporal derivative within a sufficiently short interval, where the velocity variation is limited. As a result, the corresponding events exhibit consistent motion characteristics. Traditional spatio-temporal graphs model events through pairwise connections, which capture local interactions but struggle to represent globally consistent motion patterns, resulting in redundant representations and limited structural modeling. Hypergraphs, as a generalization of graphs, allow hyperedges to connect any number of vertices simultaneously. This enables a direct association of event sets generated by the same motion into a single hyperedge, capturing higher-order motion associations more accurately and providing a more natural and efficient structured representation. The prerequisite for constructing the hypergraph is to estimate event velocities. Initially, we perform local spatio-temporal plane fitting to achieve a rapid preliminary estimation of the event velocities by Local Total Least Squares (LTLS) \citep{LTS}, followed by further refinement of the velocity estimates based on Markov Random Field (MRF). Finally, based on the optimized velocities, motion consistency clustering is performed to build the motion-aware hypergraph.

Initially, we perform local spatio-temporal plane fitting using Local Total Least Squares (LTLS) \citep{LTS} to obtain a rapid preliminary estimate of event velocities, followed by further refinement using a Markov Random Field (MRF).

Over short time intervals, when an object moves, the events generated along its edge form an approximate plane in the 3D spatio-temporal space. The direction of the plane's normal vector reflects the motion information of the events. Given the 3D coordinates of event positions, \(\mathbf{X} = [[x_1, y_1, t_1], [x_2, y_2, t_2], \dots, [x_n, y_n, t_n]]\), where \(n\) is the number of events, we employ the Singular Value Decomposition (SVD) to perform the fitting. And the equation of the fitted plane is:
\begin{equation}
  ax + by + ct + d = 0,   
\end{equation}
where $\small (a, b, c)$ is the normal vector of the plane obtained through SVD. Based on the vector, the initial velocity of event $i$ is estimated by: 
\begin{equation}
    \mathbf{v}_i^0 = \left[ v_x, v_y \right],
\end{equation}
where the velocity components can be derived as:
\begin{equation}
    \mathbf v_x = \frac{-a}{c},\qquad v_y = \frac{-b}{c}.
\end{equation}
 Although the LTLS method provides an initial velocity estimate, the estimation results can be unstable and inaccurate due to local fitting errors and noise.

Based on the Markov assumption, physical motion evolves continuously rather than undergoing random jumps, which aligns well with the spatio-temporal continuity of the event streams. Therefore, to refine the coarse velocity estimates $\mathbf{V}^0 = \{\mathbf{v}_1^0, \dots, \mathbf{v}_n^0\}$ derived from LTLS, we present a MRF-based optimization framework to enhance the motion consistency among events. Specifically, we seek an optimal velocity field $\mathbf{V}$ by minimizing the global energy function $E(\mathbf{V})$:
\begin{equation}
    E(\mathbf{V}) = \sum_{i} \underbrace{\|\mathbf{v}_i - \mathbf{v}_i^0\|^2}_{E_{data}} + \lambda \sum_{i,j} \underbrace{w_{ij}\|\mathbf{v}_i - \mathbf{v}_j\|^2}_{E_{smooth}},
\end{equation}
where $j \in \mathcal{N}_i$ denotes an event within the neighborhood of the current event $i$, and $w_{ij}$ is the spatio-temporal distance weight between event $i$ and event $j$. The data term $E_{data}$ serves as a fidelity constraint, ensuring that the optimized vector field remains anchored to the motion cues derived from the LTLS estimation. The smoothing term $E_{smooth}$ encapsulates the motion consistency assumption. By modeling the strong correlation between event $i$ and its neighborhood, this term imposes a high energy penalty on the random jumps caused by noisy data, effectively suppressing random noise that violates the principle of physical continuity. By minimizing the energy function, we can obtain the optimized velocity:
\begin{equation}
    \mathbf{v}_i = (1 - \lambda) \mathbf{v}_i^0 + \lambda \sum_{j} w_{ij} \mathbf{v}_j,
\end{equation}
where $\lambda$ is the smoothing coefficient, which balances the influence between the initial velocity estimate and the consistency of its neighboring motion patterns. This optimization mechanism promotes the generation of coherent velocity fields, laying a solid foundation for constructing robust motion-consistent hyperedges.

Finally, each event is treated as a node in the hypergraph; for instance, \(e_i\) corresponds to the \(i\)-th node in the hypergraph. And as shown in Fig. \ref{image:3}, we connect these events to form hyperedges based on the optimized event velocities, thereby constructing the hypergraph $\mathcal{G}^{{H}} = (\mathcal{V}^{H}, \mathcal{E}^{H})$. Here, \(\mathcal{V}^{H}\) denotes the set of nodes and \(\mathcal{E}^H\) denotes the set of hyperedges, where each hyperedge represents a cluster of events sharing a consistent motion pattern. Assuming we generate \(K\) clusters based on event velocities: \(C_1, C_2, \dots, C_K\), then the hyperedge set is given by \(\mathcal{E}^H = \{\mathcal{E}_k^H\}_{k=1}^K\), where \(\mathcal{E}_k^H= \{e_i \mid e_i \in C_k\}\).

\subsection{Hierarchical Hyperbolic Perception}

Although existing GNN-based methods have achieved remarkable success in event-based tasks, event stream perception in traditional pure Euclidean geometric space still struggles to effectively represent the complex and hierarchical structure inherent in event data. Thus, we propose EHGCN which first introduces HGCN for event perception. In EHGCN, hyperbolic space, with its exponential embedding nature, can encode hierarchical or fractal structures within limited dimensions, aligning seamlessly with the spatio-temporal evolution of event streams. Additionally, the negative curvature of hyperbolic space preserves geometric correlations among distant events, effectively capturing cross-region and cross-temporal event dependencies and modeling long-range interactions.

\textbf{Dual-Space Interaction. }To enable feature interaction between Euclidean and hyperbolic spaces in our EHGCN approach, we perform exponential and logarithmic maps. Given Euclidean features $\small\ \mathcal{Z}\in \mathrm{R}^n$ of events, we embed them into hyperbolic space via exponential mapping $\small \mathcal{H}=\exp _{\mathbf{o}}^{\mathrm{c}}\left( \mathcal{Z} \right)$, where $\small \mathcal{H}$ is the feature of the corresponding event in the hyperbolic space. And the computation formula is:
\begin{equation}
\small \exp _{\mathbf{o}}^{\mathrm{c}}\left( \mathcal{Z} \right) =\mathbf{o}\oplus _c\left( \tanh \left( \sqrt{\mathbf{c}}\frac{\left\| \mathcal{Z} \right\|}{2} \right) \frac{\mathcal{Z}}{\sqrt{\mathbf{c}}\left\| \mathcal{Z} \right\|} \right), 
\end{equation}
where the $\small\mathbf{o}$ denotes the origin in the Poincaré ball model, and $\oplus _c$ is the Möbius addition. Likewise, the logarithmic map performs the inverse operation, projecting features from the hyperbolic space back into the tangent space $\small \mathbf{T}_{\mathbf{o}}\mathrm{D}_{c}^{n}$. In hyperbolic space, cross-convolutional layer features for events can be fused using Möbius addition: $\small\mathcal{H} _{i}=\mathcal{H} _{i}\oplus _c\sum_{j\in N\left( i \right)}{\mathcal{H} _{j}}$, where $\small N\left( i \right) $ denotes the neighbors of event $e_i$.

\begin{figure*}[t]
  \centering
   \includegraphics[width=1\linewidth]{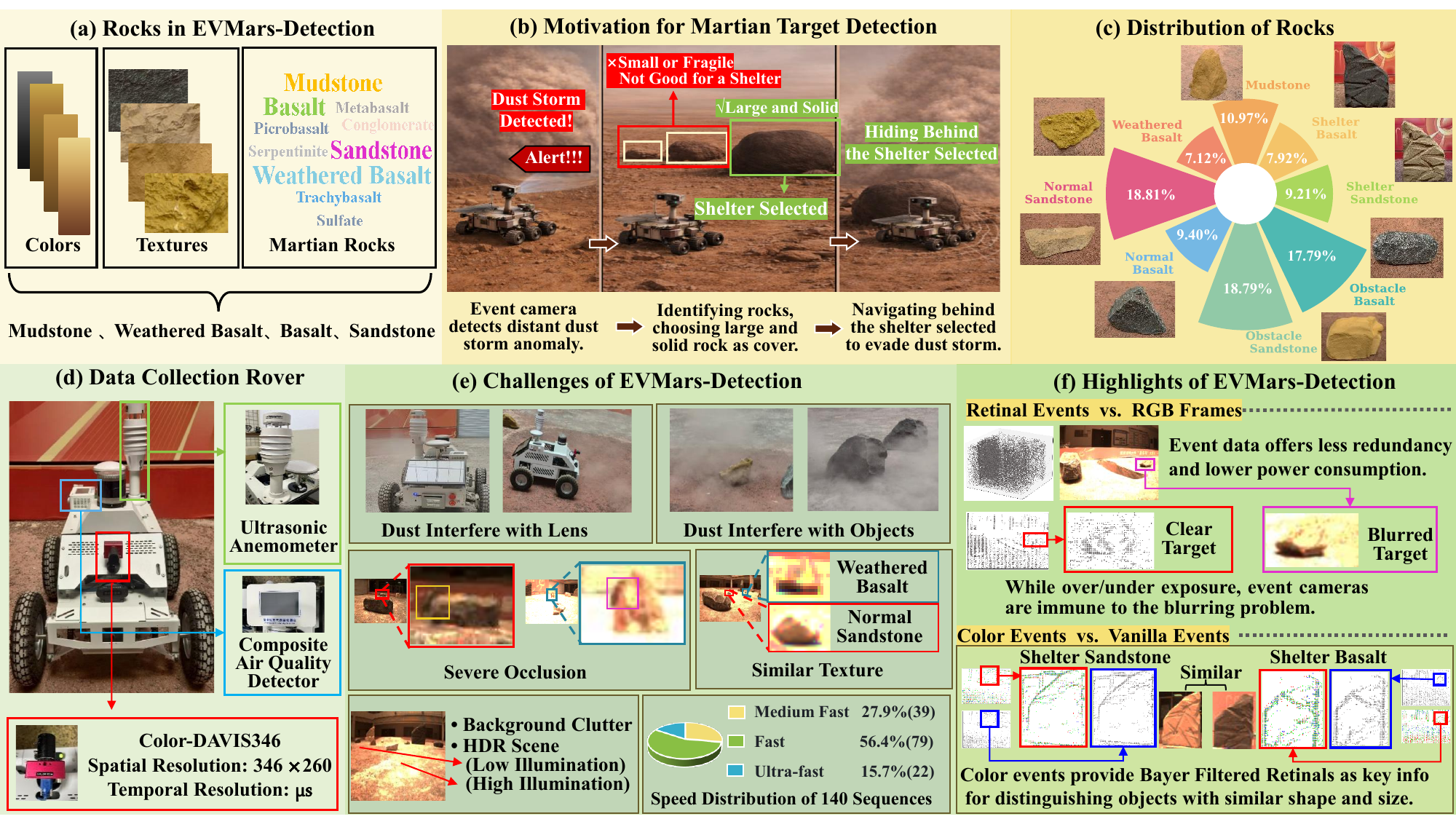}
  
   \caption{Introduction of our EVMars-Detection dataset.}
    
   \label{image:emars}
\end{figure*}

\textbf{Hierarchical Event Hyperbolic Embedding (HEHE) Block.} Let the input feature at layer $\ l$ be $\small\ {\mathcal{H} _{i}}^l\in \mathrm{D}_{c}^{n}$, as shown in Fig. \ref{image:1}, the message passing process of a HEHE block can be divided into three steps. Event-based hyperbolic feature transformation is the first step. It includes hyperbolic linear transformation and bias addition:
\begin{equation}
    \small{\widetilde{\mathcal{H} }_{i}}^l=\mathbf{W}\otimes _{c^{\mathrm{in}}}{\mathcal{H} _{i}}^l\oplus _{c^{\mathrm{in}}}\mathbf{b},
\end{equation}
where $\small \mathbf{W}\in \mathcal{R}^{d^{\prime}\times d}$ is the weight matrix, $\small \mathbf{b}\in \mathbf{T}_{\mathbf{O}}\mathrm{D}_{\mathrm{c}}^{\mathrm{n}}$ is the bias vector defined on the tangent space at the origin ,$\otimes _{c^{\mathrm{in}}}$ denotes  Möbius Matrix-Vector Multiplication and $\small c^{\mathrm{in}}$ denotes the input curvature. The hyperbolic feature transformation leverages the Euclidean isomorphism of the tangent space, directly reusing mature linear layer and bias operations. This avoids the need to redesign hyperbolic convolutional kernels. It maps node embeddings from the previous layer to the next layer, captures information from larger neighborhoods, and crucially preserves the hyperbolic geometric structure. This enhances the model's expressive power for complex relationships.

The second step is hyperbolic aggregation. It captures both local and global dependencies in event data by performing weighted aggregation of neighboring features through message passing, which can be expressed as:
\begin{equation}
    \small {\hat{\mathcal{H}}_{i}}^l=\exp _{\mathbf{o}}^{c^{\mathrm{in}}}\left( \sum_{j\in N\left( i \right)}{A_{i,j}} \log _{\mathbf{o}}^{c^{\mathrm{in}}}\left( {\widetilde{\mathcal{H} }_{j}}^l \right) \right).
\end{equation}

In tangent space, the transformed outputs of all neighboring nodes $\small {\widetilde{\mathcal{H} }_{j}}^l$ are weighted and aggregated via the adjacency matrix $A_{i,j}$, and then mapped back to hyperbolic space, narrowing the hyperbolic distance among events with analogous spatio-temporal features, substantially boosting event-based motion-consistent association.

Hyperbolic activation is the final step. While preserving the hyperbolic structure, nonlinear transformations are applied to boost expressiveness, enabling the model to learn complex patterns such as classification boundaries and hierarchical relationships. It can be expressed as follows:
\begin{equation}
    \small {\mathcal{H} _{i}}^{l+1}=\exp _{\mathbf{o}}^{c^{\mathrm{out}}}\varphi \left( \log _{\mathbf{o}}^{c^{\mathrm{in}}}\left( {\hat{\mathcal{H}}_{i}}^l \right) \right),
\end{equation}
where $\small \varphi \left( \cdot \right) $ denotes the activation function, and $\small c^{\mathrm{out}}$ is the output curvature that will serve as the input curvature for the next layer of HGCN. By combining Euclidean activations in the tangent space with exponential mapping back to the manifold, this approach also avoids the numerical instabilities of taking derivatives directly on the hyperbolic space.

Moreover, the curvature parameter $\ c$ critically governs the hyperbolic geometry all nonlinear operations. Then, $\ c$ is formulated as a learnable parameter optimized via gradient descent in this work: $\small  c^{l+1}=c^l-\eta \partial \mathcal{L}/{\partial c^l}$, where the $\small\mathcal{L}$ is the overall loss function. This adaptive curvature mechanism can dynamically adjust the hyperbolic curvature based on the characteristics of the event stream, enabling the network to adaptively select a more appropriate geometry at different scales and align the hyperbolic space with the intrinsic geometry of the event data, thereby reducing embedding distortion and enhancing geometric representation fidelity for event perception tasks and spatio-temporal association.

\section{Experiments}

\subsection{Datasets}

We evaluate the perception performance of our model on four public and representative event-based vision datasets, along with our newly constructed Mars-like event streams detection dataset, namely EVMars-Detection. Specifically, recognition tasks are conducted on the N-Cars and DVS128 Gesture datasets, while object detection tasks are evaluated on the N-Caltech101, Gen1, and our EVMars-Detection datasets.

\subsubsection{Public Event-based Datasets}
We use four widely adopted public event-based datasets, including N-Cars, DVS128 Gesture, N-Caltech101, and Gen1, to evaluate the proposed approach across representative recognition and detection scenarios involving driving scenes, indoor gestures, converted image-based events, and large-scale open road scenes.

N-Cars \citep{ncars}: An event dataset recorded in real-world driving scenarios under natural lighting and motion conditions, focusing on vehicle-background discrimination.

DVS128 Gesture \citep{dvs128}: A real-world event dataset comprising 11 hand gesture categories collected from 29 subjects under three lighting conditions.

N-Caltech101 \citep{ncaltech101}: Covers 101 object categories, with 40 to 800 samples per class, generated by converting the Caltech101 image dataset into event streams.

Gen1 \citep{gen1}: The dataset contains 39 hours of open-road driving experience, covering various driving scenarios such as urban, highway, suburban, and rural areas. It currently provides manually annotated bounding boxes for two object categories: pedestrians and cars.

\subsubsection{EVMars-Detection Dataset}

To further validate the effectiveness and generalization capability of our approach beyond conventional event-based perception scenarios, we construct EVMars-Detection. It introduces planetary exploration scenarios, providing a tough benchmark for examining stable perception performance in complex extraterrestrial environments.

\textbf{Dataset Construction and Category Design. }Existing event-based datasets related to planetary environments mainly target SLAM, navigation, or visual odometry \citep{Mars_data}, making them difficult to use for validating our approach. Thus, we construct the EVMars-Detection dataset. As shown in Fig. \ref{image:emars}(d), the dataset is collected using a mobile Mars-analog acquisition platform equipped with a Color-DAVIS346 color event camera. During data acquisition, dust disturbance is introduced, while wind and air-quality sensors are used to record environmental conditions, simulating dust interference and complex perception conditions that may occur on the Martian surface. As shown in Fig. \ref{image:emars}(a) and Fig. \ref{image:emars}(b), the dataset selects four common rock types on the Martian surface and defines eight categories according to the practical requirements of rover autonomous perception, including obstacle avoidance and potential shelter selection. Fig. \ref{image:emars}(c) shows the sample distribution of these categories.

\begin{figure*}[t]
  \centering
   \includegraphics[width=0.99\linewidth]{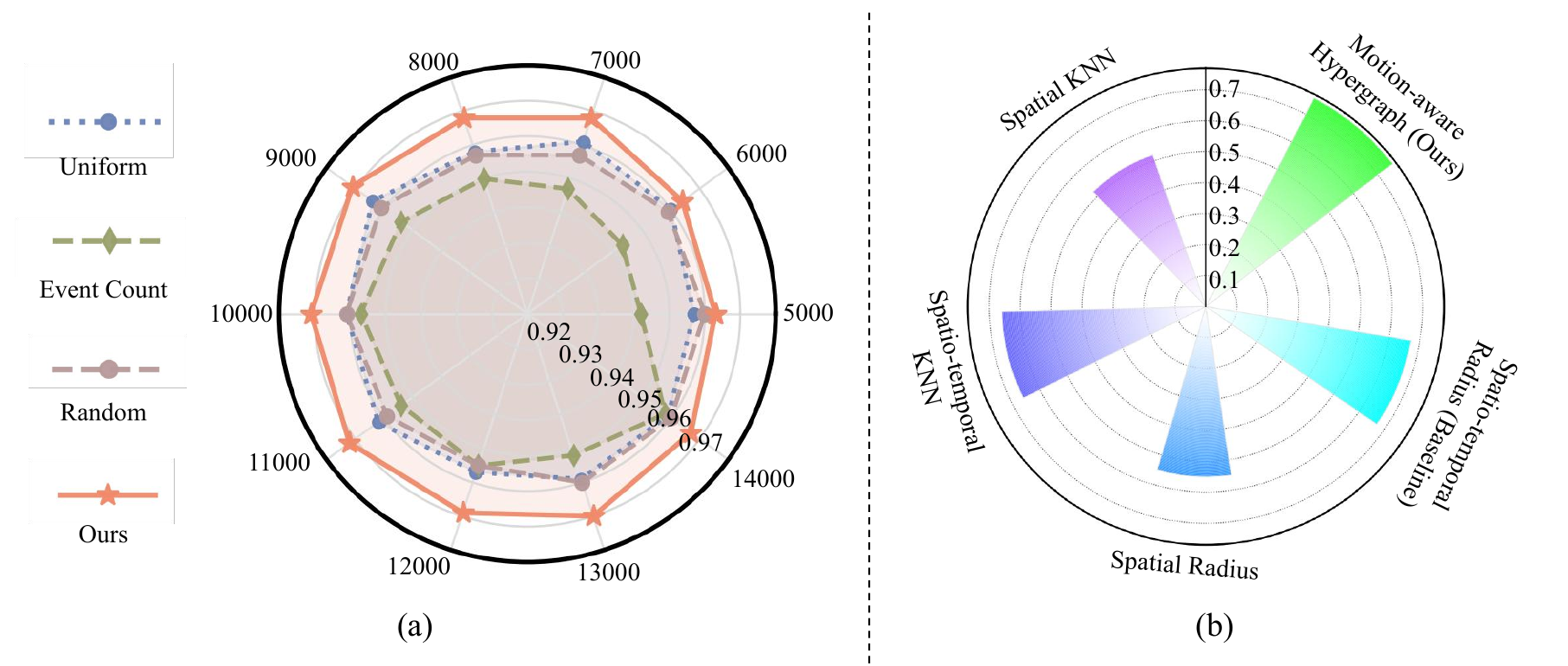}
   \caption{In the N-Cars recognition task, we compared the recognition accuracy under different sampling budgets, as shown in (a). In the N-Caltech101 detection task, we further compared graph construction methods, as shown in (b). The overall results across both datasets and tasks consistently demonstrate that our method achieves higher performance, greater stability, and better robustness to scene variations, reflecting clear mechanistic advantages.}
   \label{image:789}
\end{figure*}

\textbf{Challenges.} As shown in Fig. \ref{image:emars}(e), EVMars-Detection includes several challenging factors. Dust may interfere with the camera lens or occlude rock targets, leading to incomplete event responses. Inter-class texture and color similarity also increases fine-grained discrimination difficulty. In addition, the dataset contains severe occlusion, complex backgrounds, and HDR illumination scenes. According to the motion statistics, it covers Medium Fast, Fast, and Ultra-fast sequences, enabling robustness evaluation under multi-velocity motion distributions.

 \textbf{Highlights.} As shown in Fig. \ref{image:emars}(f), Compared with conventional RGB frames, event cameras offer low redundancy, low power consumption, and HDR, making them suitable for resource-constrained planetary robots. Under over-exposure or under-exposure conditions, event cameras can better preserve object boundaries and reduce target blurring caused by frame saturation. Moreover, color events provide additional chromatic cues for distinguishing rocks with similar shapes and sizes but different materials, such as Shelter Sandstone and Shelter Basalt. Therefore, EVMars-Detection not only introduces realistic detection challenges in complex environments, but also highlights the potential of color event cameras for planetary terrain perception.

\begin{table}[t]
\centering
\caption{Ablation study on N-Cars, where \ding{51} indicates the adoption of the corresponding component.}
\begin{tabular}{cccc}
 \midrule
 Sifting &Motion-Aware &Hyperbolic &Recognition \\
(DAES) &Hypergraph &Embedding  &Accuracy\\
 \midrule
\ding{55}    & \ding{55}    &\ding{55}     &0.945 \\
\ding{51}   &\ding{55}   &\ding{55}  &0.951  \\
\ding{55} &\ding{55}   &\ding{51}   &0.954 \\
\ding{55}   &\ding{51}  &\ding{51}  &0.959 \\
\ding{51}    &\ding{55} &\ding{51}     &0.962  \\
\ding{51}     &\ding{51}  &\ding{51}    &\textbf{0.971} \\

 \midrule
\end{tabular}

\label{table:3}
\end{table}

\subsection{Ablation}

To evaluate the contribution of the key components in EHGCN, we analyze the effectiveness of our Distribution-Aware Event Sifting (DAES) method, motion-aware hypergraph and hyperbolic embedding. The experiments are conducted on the N-Cars dataset, and Tab.\ref{table:3} presents the recognition accuracy for different combinations of components.

Notably, if DAES is not used, uniform sampling is applied. And if motion-aware hypergraphs are not used, the model defaults to a representative conventional pairwise-connected spatio-temporal graph proposed by \citet{aegnn}. In addition, the \textit{Hyperbolic Embedding} in this table indicates Euclidean–hyperbolic fusion. So its absence refers to the use of the baseline method (i.e., AEGNN). Compared to uniform sampling, our DAES method alone leads to a 0.6\%–0.8\% accuracy improvement (row 1 vs. row 2 and column 3 vs. column 5). Notably, the sole integration of hyperbolic geometry results in approximately 1\% accuracy improvement (row 1 vs. row 3), and with the addition of the hypergraph structure, an additional ~0.4\% gain is observed (row 3 vs. row 4). When all three components are integrated, the model achieves the best overall performance, further indicating that they can complement each other and fully leverage their respective advantages through synergistic effects.

Specifically, DAES improves data quality by retaining more informative events in dynamic regions and suppressing chaotic noise in static regions, thus exhibiting superior robustness in scenes with rapid motion or uneven event distribution. The MRF-optimized motion-aware hypergraph effectively captures spatio-temporal dependencies among events, enhancing global motion consistency while suppressing cross-target spurious associations, which is critical for precise detection in complex dynamic scenes. Moreover, hyperbolic embedding enhances the model’s ability to capture long-range dependencies and hierarchical feature, enabling it to handle more complex spatio-temporal structures. With the effect of all three components, the model becomes more robust in handling dynamic and complex scenarios, significantly improving accuracy. 

\begin{figure*}[t]
  \centering
   \includegraphics[width=1\linewidth]{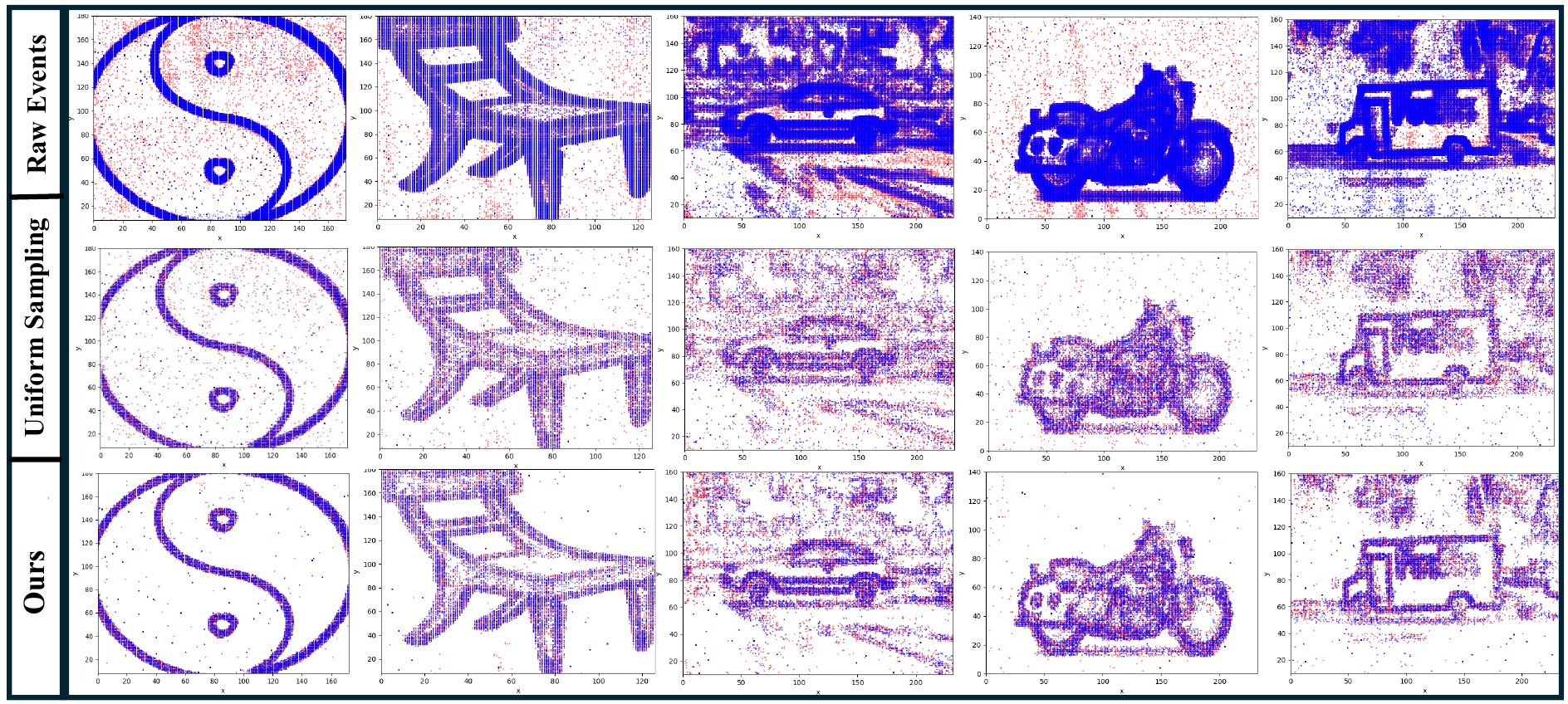}
   
   \caption{Visual comparison of events after sampling on the N-Caltech101 dataset. Compared to widely used uniform sampling method, our approach preserves important details, such as the contours of objects and motion information, while suppressing background noise and redundant information.
}
   \label{image:4}
\end{figure*}

With all other settings kept identical, we evaluate the effectiveness of the proposed DAES method against common sampling method, and investigate whether the motion-aware hypergraph provides advantages over conventional pairwise graph construction. Consistent results across two datasets and two tasks demonstrate that our method not only achieves superior performance but also exhibits enhanced stability and robustness under diverse scene variations, validating the effectiveness of the proposed event relation modeling method.

On the N-Cars recognition task, we compare accuracy under different sampling budgets as shown in Fig. \ref{image:789}(a). \textit{Uniform} follows the baseline scheme, while \textit{Event Count} and \textit{Random} are strategies reported by \citet{6sample}. Across the full range from 5000 to 14000 sampled events, our DAES consistently achieves the best performance, with stable gains rather than sporadic improvements. Under a fixed sampling budgets, the true bottleneck for accuracy lies not in how many events are retained, but in how many events with higher information content are preserved. \textit{Uniform} sampling spreads selections globally, which tends to keep redundant triggers in low information regions while discarding critical structures in fast motion or edge dense areas. \textit{Random} sampling exhibits higher variance and sensitivity to randomness, leading to more unstable results as the budget changes. \textit{Event Count} favors dense regions, yet density does not equate to informativeness, making it susceptible to background noise or local over-triggering. In contrast, our DAES method balances local details and global structure across scales, retaining more discriminative events in dynamic regions while suppressing chaotic noise in static regions. This yields higher accuracy and stronger robustness to budget variation and scene changes.

On the N-Caltech101 detection task, we further compare different graph construction methods as shown in Fig. \ref{image:789}(b). The Baseline \textit{Spatio-temporal Radius} uses spatial-temporal distance metrics to build pairwise graphs under fixed radius and maximum neighbor constraints. \textit{Spatial Radius} performs spatial neighbor search within a fixed time window while maintaining the same maximum neighbors and radius as the Baseline. \textit{Spatial KNN} and \textit{Spatio-temporal KNN} perform K-nearest neighbor connections based on spatial distance under the same time window setting or spatio-temporal distance respectively, with K aligned to the baseline's maximum neighbor number for fair comparison. The radius-based or KNN methods essentially model local pairwise relationships, which are susceptible to factors such as event density and speed variations, leading to incorrect associations across objects, local structure fractures, or semantically mixed neighborhoods. Particularly in event streams characterized by sparsity and high non-uniformity, fixed radius suffers from over-connection in dense regions and under-connection in sparse regions, while KNN may force connections between distant points in sparse regions, increasing the risk of incorrect associations. Our motion-aware hypergraph goes beyond mere pairwise connectivity. It constructs hyperedges based on motion consistency, suppressing false associations across targets. This approach preserves spatio-temporal consistency within targets and reduces interference from background and non-target events in complex dynamic scenes. By shifting from local pairwise relationships to group-level consistency, our scheme achieves better detection performance. 

\begin{figure}[t]
    \centering
    \includegraphics[width=1\linewidth]{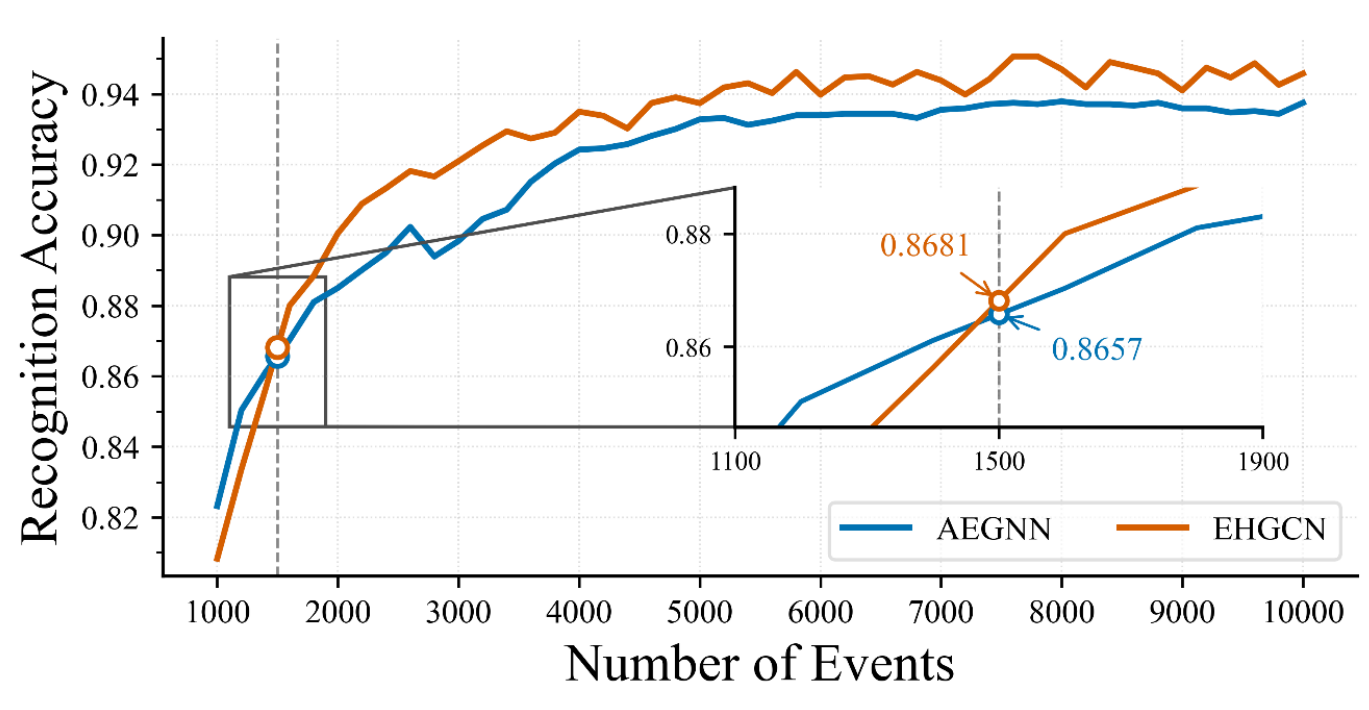}
    \caption{Accuracy over events. Our approach surpasses AEGNN at around 1,500 events and reaches over 90\% accuracy with approximately 2,000 events.} 
    \label{figure:5}
\end{figure}

\begin{figure*}[t]
  \centering
   \includegraphics[width=1\linewidth]{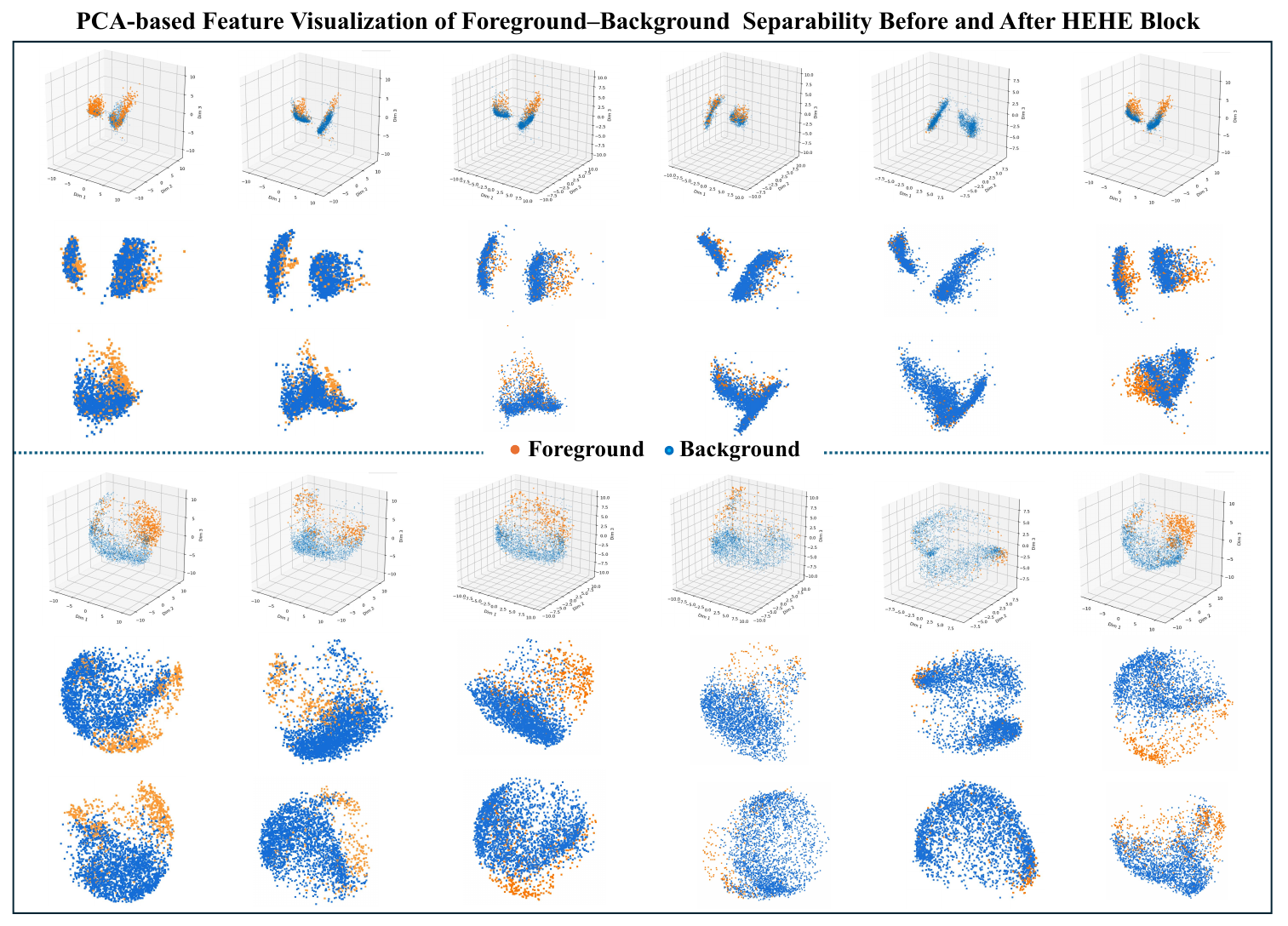}
 
   \caption{Feature Distribution Visualization from PCA Perspective. PCA visualizes the feature distributions of foreground and background nodes before and after the HEHE block from the perspective of linear projection. The upper part presents the feature distributions before HEHE processing, while the lower part shows the corresponding distributions after HEHE processing. Each column represents the same sample, and three different viewpoints are provided for each sample to comprehensively illustrate the feature distribution changes in the projected space.}
   \label{image:pca_vis}
\end{figure*}

\begin{figure*}[t]
  \centering
   \includegraphics[width=1\linewidth]{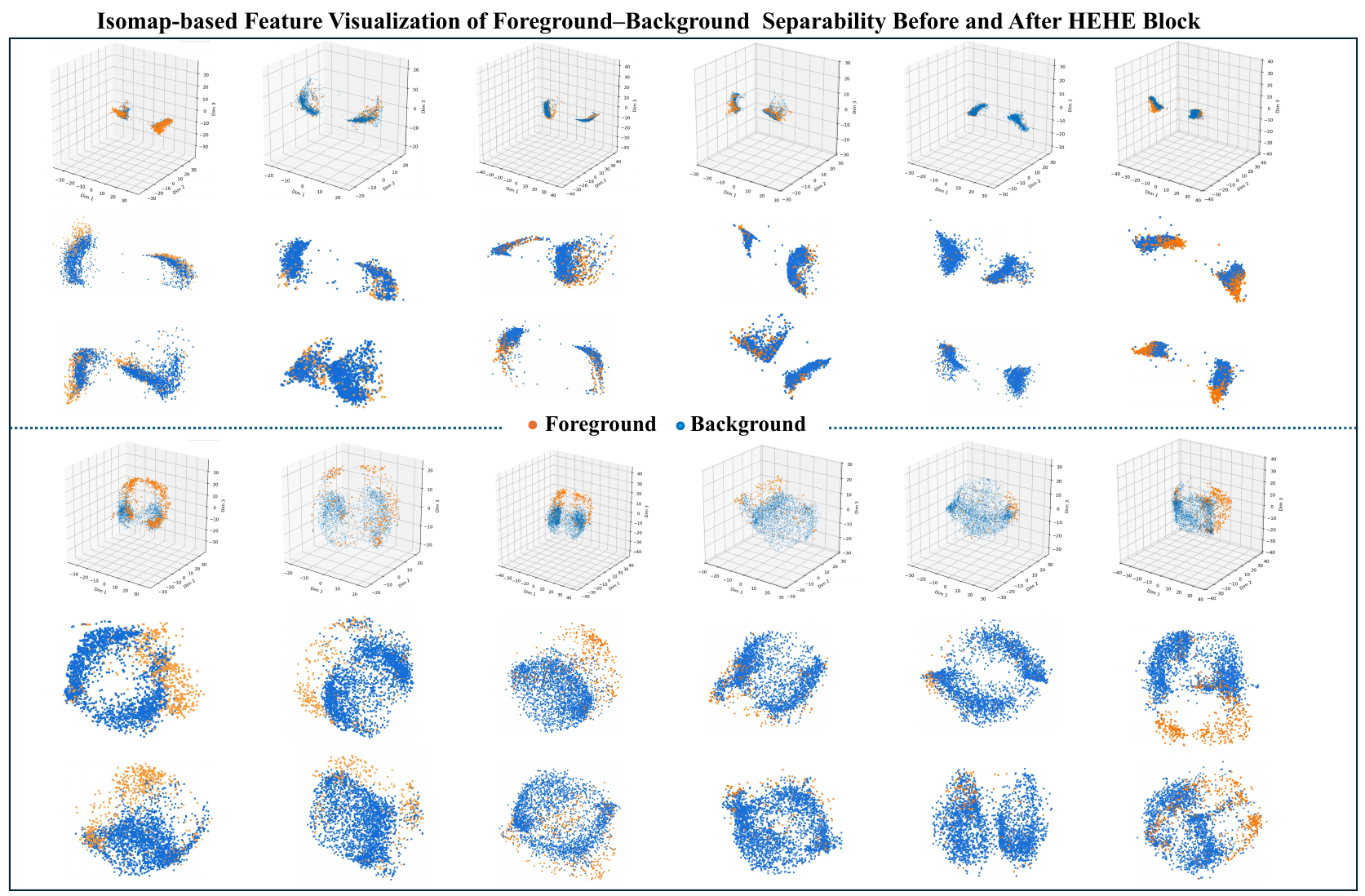}
  
   \caption{Feature Distribution Visualization from Isomap Perspective. Isomap visualizes the foreground and background feature distributions before and after the HEHE block from the perspective of nonlinear manifold learning. Similar to the PCA-based visualization, after HEHE processing, foreground nodes maintain strong local compactness in the low-dimensional manifold space, while background nodes occupy a broader manifold region.}
   \label{image:separability}
\end{figure*}
Moreover, in the ablation study, we conduct a qualitative evaluation of DAES on the N-Caltech101 dataset. As shown in Fig. \ref{image:4}, although the raw events are densely distributed, they contain a significant amount of noise. In contrast, the baseline method of \citet{aegnn} employs uniform sampling, which is susceptible to substantial noise interference and information loss, whereas our DAES effectively suppresses noise, yielding cleaner and structurally clearer result. Specifically, our approach, using a multi-scale distribution-aware event sifting method, effectively filters out outlier noise, adjusts for event density in different regions, and preserves important details such as object contours and motion information, while suppressing background noise and redundant data. This enables our method to maintain higher visual clarity and object recognition while improving information density.

Our model does not require the full event stream used during training to make correct predictions. To further evaluate its adaptability to varying event counts, we measure the classification accuracy on the N-Cars dataset across different numbers of events and systematically compare the results with the representative AEGNN method. As shown in Fig. \ref{figure:5}, our approach surpasses AEGNN at around 1,500 events and reaches over 94$\%$ accuracy at about 5,400 events, which shows the feasibility and superiority of our approach for event-based vision perception tasks.

\begin{figure*}[t]
  \centering
   \includegraphics[width=0.99\linewidth]{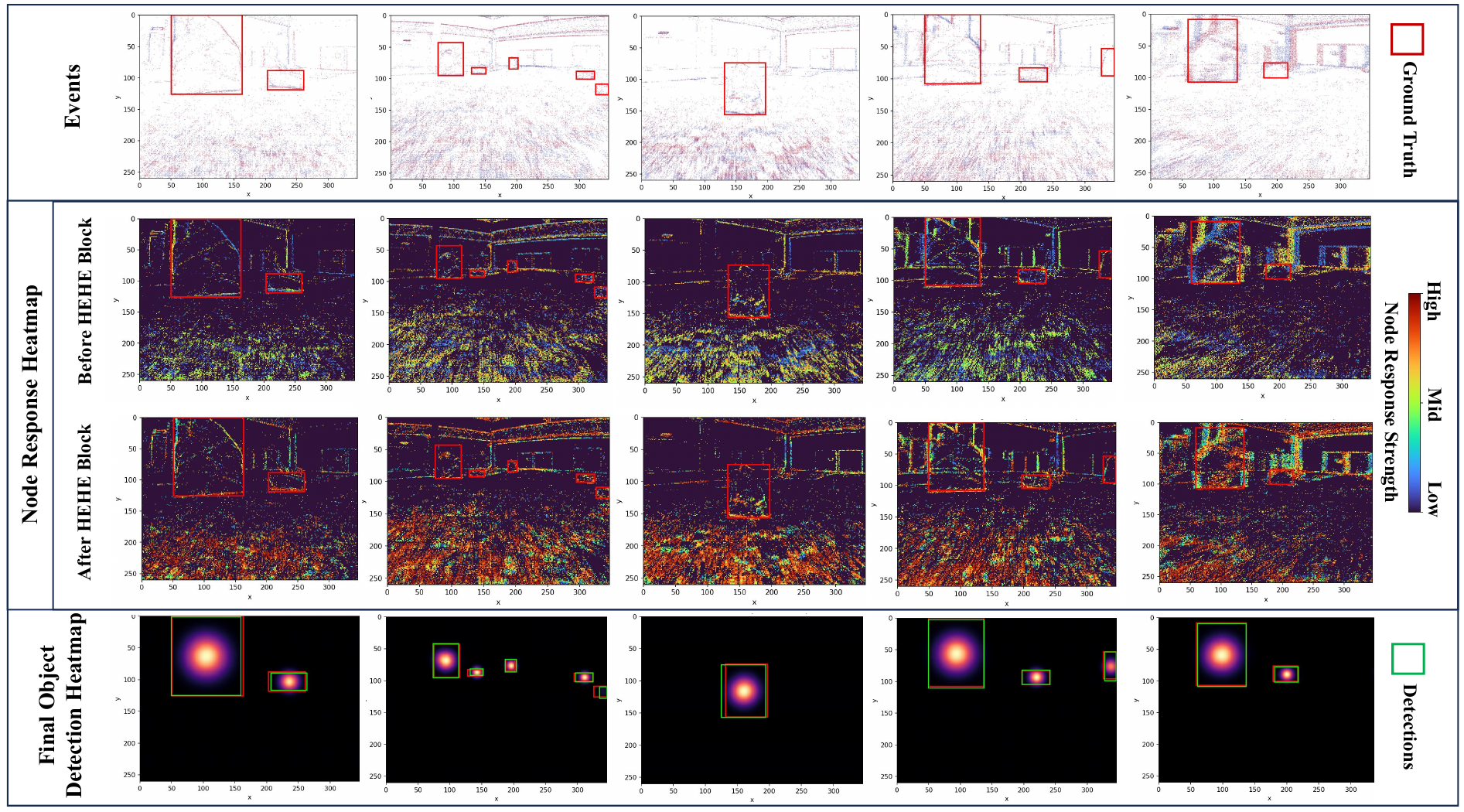}

   \caption{Visualization of node responses before and after hyperbolic embedding and the final detection responses. The first row shows sampled event inputs, with red boxes indicating ground-truth annotations. The second and third rows show node response heatmaps before and after HEHE block, while the last row presents final detection responses with green predicted boxes. Before hyperbolic embedding, node responses are relatively smooth and mainly capture local edges and background textures. After HEHE processing, the responses exhibit clearer hierarchical patterns, with stronger activations around object contours, structural boundaries, and event-dense regions.  }
   \label{image:heatmap}
\end{figure*}

\begin{table*}[t]
\centering

\caption{Recognition accuracy on N-Cars and DVS128.}

\label{table:1}



\begin{tabular}{@{}lcccc@{}}
\toprule
Method & Representation & N-Cars $\uparrow$ & DVS128 $\uparrow$ & GFLOPs $\downarrow$ \\
\midrule

RG-CNNs \citeyearpar{RGCNNS_recog} 
& Point-based & 0.914 & 0.972 & 0.79 \\

EST \citeyearpar{EST_recog} 
& Frame-based & 0.919 & - & 4.28 \\

M-LSTM \citeyearpar{re_classification_CNN} 
& Frame-based & 0.927 & - & 4.82 \\

MVF-Net \citeyearpar{MVFNET_recog} 
& Frame-based & 0.927 & - & 5.62 \\

VMV-GCN \citeyearpar{VMvgcn_recog} 
& Point-based & 0.932 & 0.975 & 1.30 \\

AEGNN \citeyearpar{aegnn} 
& Point-based & 0.945 & - & 0.75 \\

EV-VGCNN \citeyearpar{EVVGCNN_recog} 
& Point-based & 0.953 & 0.959 & 0.70 \\

ECSNet-SES \citeyearpar{ECSNe_recog} 
& Point-based & 0.946 & 0.986 & 1.83 \\

VMST-Net \citeyearpar{VMST_recog} 
& Point-based & 0.944 & 0.978 & 0.14 \\

GET \citeyearpar{get_recog} 
& Frame-based & 0.967 & 0.979 & 3.10 \\

EDGCN \citeyearpar{EDGCN_recog} 
& Point-based & 0.958 & 0.985 & 0.57 \\

SpikMamba \citeyearpar{spikmamba} 
& Point-based & - & 0.990 & 0.12 \\

EVA \citeyearpar{EVA_recog} 
& Frame-based & 0.963 & 0.969 & - \\

EventMamba \citeyearpar{eventmamba} 
& Point-based & - & 0.992 & 0.22 \\

EHGCN (Ours) 
& Point-based 
& \textbf{0.971} 
& \textbf{0.993} 
& \textbf{0.11} \\

\bottomrule
\end{tabular}


\end{table*}

\subsection{Hyperbolic Geometric Visualization}
To further investigate the working mechanism of the HEHE block, we conduct a progressive visualization analysis from four perspectives: global feature geometry, nonlinear manifold structure, local node responses, and final detection responses. Event streams are inherently sparse, non-uniform, and hierarchically structured. In event graphs, object contours, rock boundaries, locally dense events, background textures, and noise responses jointly form complex structural relationships. Although local graph convolution in Euclidean space can aggregate neighboring event information, its capacity to represent hierarchical structures is relatively limited. Consequently, structural patterns with different semantic meanings or hierarchical levels may be compressed into adjacent feature regions, resulting in entangled representations of target-related and complex background nodes. Motivated by this limitation, we introduce hyperbolic geometry to model the hierarchical and non-Euclidean relationships in event graphs and analyze its effect through the following visualization experiments.

First, event nodes are categorized as foreground or background according to whether they lie inside the ground-truth bounding boxes. PCA is then employed to compare their feature distributions before and after the HEHE block from the perspective of linear projection. As shown in Fig. \ref{image:pca_vis}, before entering the HEHE block, foreground and background nodes exhibit considerable overlap in the low-dimensional space, indicating that the Euclidean features have not yet formed a clear global discriminative structure. After the HEHE block, the two groups exhibit more distinct structural differences in the embedding space. Foreground nodes are no longer randomly scattered but tend to form relatively compact distributions along specific directions, whereas background nodes occupy broader regions. These results indicate that hyperbolic representation enhances the intra-class consistency of target-related nodes while enlarging the structural discrepancy between foreground and background features, thereby improving their separability under linear projection.

However, PCA mainly characterizes feature distributions along linear variance directions. Therefore, PCA alone is insufficient to determine whether the observed separation remains stable in nonlinear neighborhood structures. To this end, we further employ Isomap to visualize the features before and after the HEHE block from the perspective of nonlinear manifold learning. As shown in Fig. \ref{image:separability}, before HEHE processing, foreground and background nodes remain substantially overlapped in the manifold space, and their local neighborhoods do not exhibit stable class-specific structures. After the HEHE block, foreground nodes demonstrate stronger local compactness, whereas background nodes occupy broader and more dispersed manifold regions. The consistent trends observed in PCA and Isomap indicate that the improved foreground–background separation is not an incidental result of a particular linear projection direction, but is also preserved in nonlinear neighborhood and manifold structures. These observations further suggest that hyperbolic space can more effectively encode the hierarchical, non-uniform, and non-Euclidean structural relationships of event graphs.

Although PCA and Isomap reveal the effect of HEHE on foreground–background separability in the global feature space, the low-dimensional embeddings do not directly explain how such separation is formed within the event graph. We therefore further visualize the node responses before and after the HEHE block to analyze how hyperbolic representation models the internal structures of both foreground and background regions. As shown in Fig. \ref{image:heatmap}, before entering the HEHE block, the node responses are relatively smooth. Textures, edges, and event-dense regions in both foreground and background mainly exhibit similar low-to-medium response intensities, without forming a clear structural hierarchy.

After the HEHE block, more pronounced low, medium, and high-level responses emerge in both foreground and background regions. This phenomenon does not indicate that the model simply enhances foreground responses while suppressing background responses. Instead, it suggests that HEHE performs more comprehensive hierarchical modeling of foreground objects, complex backgrounds, and local structural cues within the event graph. Specifically, in foreground regions, responses with different intensities form relatively consistent spatial organizations around object contours, boundaries, and internal structures. In background regions, ground textures, background edges, local corners, and event-dense areas may also produce salient responses, but their spatial continuity, structural consistency, and contextual relationships are different from those of target-related regions.

Therefore, HEHE does not distinguish foreground from background according to a single response-intensity threshold. Instead, it learns more discriminative scene representations by modeling the hierarchical structures and organization patterns among different regions. This process enables the model to simultaneously learn and perceive foreground objects and complex background structures, and further understand more robust, stable, and trustworthy scene features, such as the corner, boundary, and local structural features shown in the Fig. \ref{image:8} and Fig. \ref{image:heatmap} . The increased compactness of foreground nodes in the PCA and Isomap results indicates that more consistent hierarchical relationships are established among target contours, textures, and local structures. In contrast, the broader distribution of background nodes shows that the model preserves and differentiates the structural diversity of complex background textures, edges, noise responses, and local event patterns.

The final detection heatmaps further demonstrate that the detection head can exploit the hierarchical structural representations learned by HEHE to concentrate responses more stably on the ground-truth target regions. Overall, HEHE first unfolds multilevel response structures within both foreground and background regions, then learns their distinct structural organization patterns, and finally transforms these hierarchical structural differences into more stable and reliable detection responses.

\begin{table*}[t]
\centering

\small
\caption{Comparison of detection performance on N-Caltech101 and Gen1.}
\label{table:2}


\centering
\setlength{\tabcolsep}{6pt}

\begin{tabular}{lcccc}
\toprule
Method & Representation & Backbone & 
N-Caltech101 $\uparrow$ & Gen1 $\uparrow$ \\
\midrule

AsyNet \citeyearpar{asynet_recog} 
& Frame-based & CNNs 
& 0.643 & 0.145 \\

RED \citeyearpar{re_detection_CNN3} 
& Frame-based & CNNs 
& - & 0.400 \\

NVS-S \citeyearpar{NVS-S} 
& Point-based & GNNs 
& 0.346 & 0.086 \\

AEGNN \citeyearpar{aegnn} 
& Point-based & GNNs 
& 0.595 & 0.163 \\

EOGNN \citeyearpar{EOGNN} 
& Point-based & GNNs 
& 0.558 & - \\

SFOD \citeyearpar{sfod} 
& Point-based & SNNs 
& - & 0.321 \\

OGRHI \citeyearpar{ogrh_recog} 
& Point-based & GNNs 
& 0.534 & - \\

EAS-SNN \citeyearpar{EASSNN_recog} 
& Point-based & SNNs 
& 0.664 & 0.409 \\

DAGr\_N \citeyearpar{EAGR_recog} 
& Point-based & GNNs 
& 0.629 & 0.263 \\

DAGr\_S \citeyearpar{EAGR_recog} 
& Point-based & GNNs 
& 0.702 & 0.304 \\

DAGr\_M \citeyearpar{EAGR_recog} 
& Point-based & GNNs 
& 0.707 & 0.318 \\

DAGr\_L \citeyearpar{EAGR_recog} 
& Point-based & GNNs 
& 0.732 & 0.321 \\

SpikSSD \citeyearpar{spikssd} 
& Point-based & SNNs 
& - & 0.408 \\

SSLA-M \citeyearpar{SSLA} 
& Point-based & Linear Attention 
& 0.724 & 0.370 \\

SSLA-L \citeyearpar{SSLA} 
& Point-based & Linear Attention 
& 0.743 & 0.375 \\

EHGCN-S (Ours) 
& Point-based & GNNs 
& 0.743 & 0.417 \\

EHGCN-L (Ours)  
& Point-based & SSM+GNNs 
& \textbf{0.759} & \textbf{0.433} \\

\bottomrule

\end{tabular}

\end{table*}

\subsection{Results}
\textbf{Implementation details.} Our Euclidean GCN employs graph convolution with B-splines \citep{B-splines}, which are piece-wise polynomial functions characterized by local support and defined in the Euclidean space (${\small\mathrm{R}^d}$). The curvature of HEHE block is treated as a learnable parameter, initialized with 0.4, while the dropout rate is set to 0.1. And we used AdamW with an initial learning rate of $10^{-3}$ for object detection and Adam with an initial learning rate of $10^{-2}$ for object recognition.

\textbf{Object Recognition.} We evaluated our EHGCN model on two benchmark event-based datasets, comparing it against advanced methods. As shown in Tab.\ref{table:1}, the frame-based and point-based data representation methods are described in detail in the related work section. The results indicate that our approach, EHGCN, directly processes raw data, effectively avoiding the information loss caused by frame discretization, and achieves superior performance across all the two datasets. Notably, our approach achieves the optimal accuracy on N-Cars (0.971) and DVS128 (0.993). While GET and EDGCN perform well in object recognition, they exhibit clear limitations in handling dynamic, sparse, and complex event data. GET relies heavily on fixed neighborhood methods for graph construction, failing to effectively adapt to dynamic changes and background interference, while EDGCN lacks sufficient dynamic adaptability in spatio-temporal modeling, leading to unstable performance in rapidly changing scenes. In contrast, EHGCN leverages motion-aware hypergraphs and distribution-aware event sifting to more accurately capture spatio-temporal relationships, enhancing robustness and accuracy in dynamic scenes. And We compare the computational cost (GFLOPs) with prominent methods on the two datasets. EHGCN requires only 0.11 GFLOPs of computational cost, achieving nearly a 30-fold reduction compared to the runner-up method GET (3.10 GFLOPs) on the NCAR dataset. And on the DVS128 dataset, its cost is also just half that of EventMamba (0.219 GFLOPs). The results show that our approach achieves the most favourable trade-off between accuracy and computational cost.

\begin{table*}[t]
\centering

\caption{Comparison of detection performance on EVMars-Detection.}
\label{tab:emars_detection}

\resizebox{\textwidth}{!}{

\begin{tabular}{lccccc}
\toprule
Method & Representation & Backbone & 
mAP $\uparrow$ & Parameter  & MFLOPs/ev $\downarrow$ \\
\midrule

AEGNN \citeyearpar{aegnn} 
& Point-based & GNNs 
& 0.271 & 20.4M & \textbf{1.21} \\

AEGNN+YOLOX-S \citeyearpar{aegnn} 
& Point-based & GNNs 
& 0.553 & 33.9M & 3.60 \\

AEGNN+YOLOX-L \citeyearpar{aegnn} 
& Point-based & GNNs 
& 0.586 & 37.7M & 3.73 \\

DAGr\_L \citeyearpar{EAGR_recog} 
& Point-based & GNNs 
& 0.366 & 8.3M & 16.10 \\

SMamba \citeyearpar{mamba} 
& Frame-based & SSM+RNN 
& 0.612 & 16.1M & 2.22 \\

EHGCN-S (Ours) 
& Point-based & GNNs 
& 0.618 & 14.1M & 3.04 \\

EHGCN-L (Ours)
& Point-based & SSM+GNNs 
& \textbf{0.649} 
& 17.9M 
& 3.17 \\

\bottomrule
\end{tabular}


}
\end{table*}

\textbf{Object Detection.} Traditional RGB-based object detection takes the captured RGB images or video frames as input to identify and locate the target categories and their positions in the 2D plane. In contrast, event-based object detection uses the asynchronous event stream generated by the event camera as input. After constructing a spatio-temporal representation, it performs recognition and localization of the target categories and their spatial positions in the scene.

To further evaluate the effectiveness of the proposed approach for event-based object detection, we conduct systematic comparisons on N-Caltech101, Gen1, and our EVMars-Detection dataset. Tab.\ref{table:2} reports the detection results on public event-based object detection benchmarks. As shown in the table, both EHGCN-S and EHGCN-L achieve competitive performance on N-Caltech101 and Gen1. Specifically, EHGCN-S, equipped with the YOLOX detection head, obtains 0.743 and 0.417 on the two datasets, respectively. Compared with EAS-SNN, which leverages event-driven spiking mechanisms to preserve temporal precision, EHGCN achieves better performance by explicitly modeling the structural correlations among asynchronous events. This indicates that relying solely on neuronal dynamics is insufficient to capture complex motion-dependent relationships and hierarchical structures within event streams. Furthermore, compared with Linear Attention-based methods (SSLA-M/L), which improve efficiency by modeling long-range dependencies through sparse sequence interactions, EHGCN further exploits the intrinsic geometric structures and hierarchical organizations of event streams. Since attention-based sequence modeling mainly focuses on feature correlations and lacks explicit topological modeling among events, it may struggle to characterize high-order motion patterns and non-uniform event distributions. Built upon EHGCN-S, EHGCN-L further incorporates an SSM-based temporal modeling module \citep{mamba}, improving the detection performance to 0.759 and 0.433, achieving the best results in the table. These results indicate that the motion-aware hypergraph structure effectively models local spatio-temporal correlations in event streams, while the SSM module further enhances long-range temporal dependency modeling, leading to improved detection stability and discriminative capability in dynamic scenes.

\begin{figure*}[t]
  \centering
   \includegraphics[width= 1\linewidth]{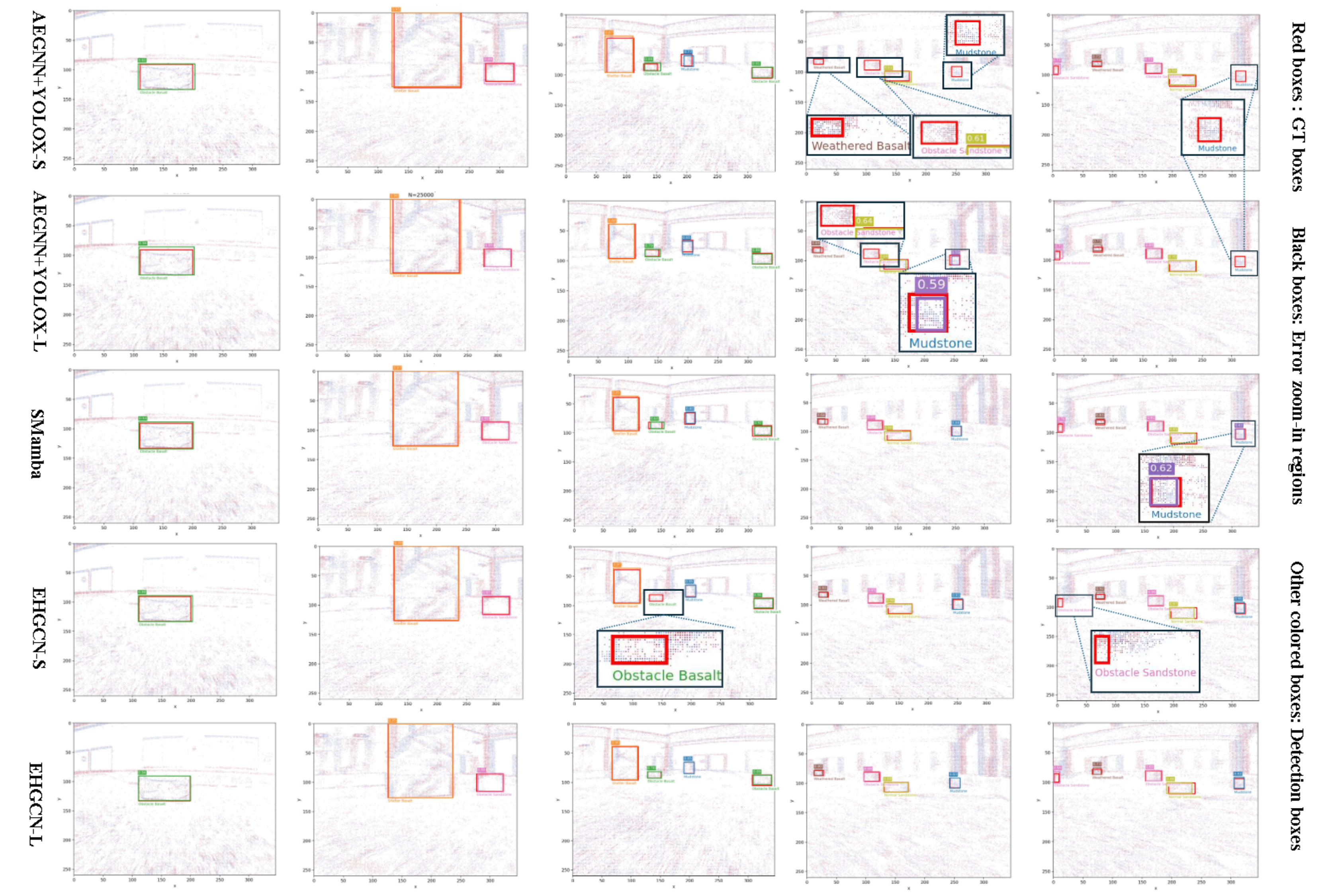}
 
   \caption{Visualization of Detection Results on EVMars-Detection. Red boxes indicate ground-truth bounding boxes, and the colored text at the lower-left corner of each red box denotes the corresponding ground-truth class label. Black boxes are used to highlight error zoom-in regions, including false detections or missed detections. The other colored boxes represent model-predicted bounding boxes, with the value at the upper-left corner indicating the confidence score of each detection. A prediction is considered correctly classified when the color of the predicted box matches the color of the corresponding ground-truth class label. }
   \label{image:emars_detection000}
\end{figure*}

On the EVMars-Detection dataset, Tab.\ref{tab:emars_detection} further compares different methods in terms of detection accuracy, model parameters, and computational cost under complex Mars-analog scenarios. For the other compared methods, we followed the original complete parameter configurations as closely as possible in data processing, model training, and related experimental settings to ensure a fair comparison. For a fair comparison at the detection head, we combine AEGNN with YOLOX-S and YOLOX-L detection heads, denoted as AEGNN+YOLOX-S and AEGNN+YOLOX-L, respectively. These two variants adopt detection heads corresponding to those used in EHGCN-S and EHGCN-L. The results show that, under the same level of detection head, EHGCN-S achieves an mAP of 0.618, outperforming AEGNN+YOLOX-S with an mAP of 0.553. Meanwhile, EHGCN-S uses only 14.1M parameters, much fewer than the 33.9M parameters of AEGNN+YOLOX-S, and reduces the computational cost from 3.60 MFLOPs/ev to 3.04 MFLOPs/ev. This demonstrates that the performance gain does not simply come from a larger detection head, but mainly benefits from the proposed hypergraph event representation and GNN-based feature modeling.

Furthermore, by incorporating the SSM module into EHGCN-S, EHGCN-L improves the mAP from 0.618 to 0.649, achieving the best detection performance on the EVMars-Detection dataset. Compared with AEGNN+YOLOX-L, which uses the corresponding detection head, EHGCN-L improves the mAP by 0.063, while reducing the number of parameters from 37.7M to 17.9M and the computational cost from 3.73 MFLOPs/ev to 3.17 MFLOPs/ev. It should be noted that such a substantial parameter reduction cannot be explained solely by differences in the detection head, but is mainly attributed to the efficient modeling capability of hyperbolic graph convolution for structural features in event streams. Compared with Euclidean graph convolution, HGCN can more effectively represent the hierarchical structures and non-Euclidean relationships in event graphs within a lower-dimensional space, thereby reducing the reliance on channel expansion and redundant parameters while maintaining favorable object localization and classification performance with lower computational cost. In addition, EHGCN-L also outperforms SMamba, improving the mAP from 0.612 to 0.649. Overall, EHGCN-L achieves a better balance among detection accuracy, parameter efficiency, and computational cost, demonstrating the complementary advantages of hypergraph representation, GNN-based feature modeling, and SSM-based temporal enhancement.

To further intuitively evaluate the detection performance of different methods in complex Mars-analog scenarios, we conduct qualitative visual comparisons on the EVMars-Detection dataset, as shown in Fig. \ref{image:emars_detection000}. It can be observed that AEGNN+YOLOX-S and AEGNN+YOLOX-L are able to detect major objects in some samples, but they still suffer from missed detections, false detections, or localization deviations in regions with sparse event responses, texture-similar backgrounds, and small objects. SMamba also shows certain detection capability in several scenes, but its discrimination of object boundaries and local structures remains less stable and can be affected by background noise and non-target event responses. In contrast, EHGCN-S can more accurately localize target regions in most samples and effectively reduce false detections in background areas, indicating that the proposed hypergraph event representation and GNN-based feature modeling enhance the discriminability of target-related structures. After further incorporating the SSM module, EHGCN-L exhibits more stable performance in terms of detection completeness, class consistency, and localization accuracy. This is particularly evident in scenes with multiple objects, weak textures, and strong background interference, where EHGCN-L can better maintain responses on target regions. The black zoom-in regions further illustrate the differences among methods in terms of false detections and missed detections, showing that EHGCN-L can effectively alleviate detection errors in complex Mars-analog environments. Overall, the qualitative results are consistent with the quantitative comparisons, further validating the robustness and effectiveness of EHGCN on the EVMars-Detection dataset.

In summary, the experimental results demonstrate that our approach not only generalizes well to public event-based detection benchmarks, but also adapts effectively to the more challenging Mars-analog scenarios in EVMars-Detection. EHGCN-S verifies the effectiveness of hypergraph event representation in suppressing background interference and enhancing target structural representation, while EHGCN-L further improves long-range temporal modeling through the SSM module, achieving a better balance between detection accuracy and efficiency.

\section{Limitation}

Despite the effectiveness of EHGCN, some limitations remain. To ensure fair comparisons with existing event-based methods, our framework mainly exploits the intrinsic spatio-temporal structures of event streams and does not explicitly incorporate the color information provided by the EVMars-Detection dataset. Although this design enables a more focused evaluation of event-based representation learning, it may leave potentially complementary color cues underutilized. Moreover, since event cameras rely on brightness changes to generate events, scenes with slow motion or near-static conditions may produce insufficient event responses, which can inevitably affect detection performance.

\vspace{-2mm}
\section{Conclusion}

In this paper, we propose a novel event perception approach called EHGCN. First, we effectively reduce chaotic noise through distribution-aware event sifting, providing a clear data foundation for subsequent perception tasks. Next, we capture motion correlations among events using a MRF-optimized motion-aware hypergraph. Finally, we propose a Euclidean-hyperbolic GCN to fuse the information densely aggregated and hierarchically modeled in local Euclidean and global hyperbolic spaces, respectively, to achieve a hybrid event perception. Experimental results on multiple public datasets and our constructed EVMars-Detection dataset demonstrate that EHGCN outperforms existing SOTA methods in event perception tasks such as object detection and recognition, while achieving a favorable balance between accuracy and computational efficiency. 

In future work, we will further explore the potential of hyperbolic geometry for event representation learning by jointly modeling color cues, spatio-temporal structures, and motion dynamics. Meanwhile, we will investigate more comprehensive hybrid perception frameworks that integrate RGB-event information and multi-modal representations, aiming to advance geometric learning-based event perception in complex real-world scenarios.

\section{Acknowledgements}
\label{sec:ack}

This work is supported by National Natural Science Foundation of China (Grants 62221005, 62472060), China Postdoctoral Science Foundation (Grant GZC20233362, 2024MD754043), Chongqing Postdoctoral Innovative Talents Support Program (Grant CQBX202316), Chongqing Municipal Education Commission (Grant KJQN202400648), and Chongqing Science and Technology Bureau (Grant CSTB2025NSCQ-GPX1261).

\section{Data Availability}
The experiments reported in this study are conducted using four publicly available event-based vision datasets and one dataset constructed in this study. Specifically, the N-Caltech101 dataset \cite{ncaltech101} is used to produce the results presented in Fig. \ref{image:789},  Fig. \ref{image:4} and Tab. \ref{table:2}. The IBM DVS128 Gesture dataset \cite{dvs128} is used for the experiments reported in Tab. \ref{table:1}. The N-Cars dataset \cite{ncars} is used to produce the results presented in Fig. \ref{image:789}, Fig. \ref{figure:5} and Tab. \ref{table:3}, Tab. \ref{table:1}. The Prophesee Gen1 Automotive Detection Dataset \cite{gen1} is used for the experiments reported in Tab. \ref{table:2}. The EVMars-Detection dataset introduced and constructed in this study is used to produce the results presented in Fig.\ref{image:pca_vis}--Fig.\ref{image:emars_detection000} and Tab. \ref{tab:emars_detection}.

N-Caltech101 is publicly available from Mendeley Data (DOI: 10.17632/cy6cvx3ryv.1). The Gen1 and N-Cars datasets can be accessed through the official Prophesee dataset resources, while the DVS128 Gesture dataset is available through the IBM Research resource associated with the original publication.

The EVMars-Detection dataset, including the event streams, annotations, category definitions, and the official training, validation, and test splits, can be accessed via the EHGCN project repository at \url{https://github.com/ev-lluo/EHGCN}. The repository also provides dataset access instructions and supporting documentation describing the data acquisition procedure, annotation protocol, category definitions, and dataset partitioning.

All third-party datasets are subject to the licences and terms of use specified by their respective data providers. The EVMars-Detection dataset and the associated materials distributed through the EHGCN project repository are subject to the licence stated in the repository.

\bibliography{bib}

\end{document}